\documentclass[letterpaper, 10 pt, conference]{ieeeconf}  

\IEEEoverridecommandlockouts                              

\usepackage{amsmath} 
\usepackage{amssymb}  

\let\labelindent\relax
\usepackage[inline]{enumitem}
\usepackage{graphicx}
\usepackage{subfigure}
\usepackage{algorithm}
\usepackage{algorithmic}

\DeclareMathOperator*{\argmax}{arg\,max}
\DeclareMathOperator*{\argmin}{arg\,min}
\DeclareMathOperator{\E}{\mathsf{E}}

\title{\LARGE \bf
Learning Skills to Navigate without a Master:
A Sequential Multi-Policy Reinforcement Learning Algorithm
}

\author{Ambedkar Dukkipati and Rajarshi Banerjee and \\
Ranga Shaarad Ayyagari  and Dhaval Parmar Udaybhai
\thanks{This work was
supported by the IMPRINT under Grant IMP/2019/000383.}
\thanks{Ambedkar Dukkipati and Ranga Shaarad Ayyagari are with the Department of Computer Science and Automation, Indian Institute of Science, Bangalore 560012, India. (e-mail: \{ambedkar,rangaa\}@iisc.ac.in).
Rajarshi Banerjee and Dhaval Parmar Udaybhai have done this work while at the Department of Computer Science and Automation, Indian Institute of Science, Bangalore 560012, India. (e-mail: \{rajarshib,dhavalparmar\}@iisc.ac.in).}
}

\begin{document}

\maketitle
\thispagestyle{empty}
\pagestyle{empty}

\begin{abstract}
Solving complex problems using reinforcement learning necessitates breaking down the problem into manageable tasks, and learning policies to solve these tasks. These policies, in turn, have to be controlled by a master policy that takes high-level decisions. Hence learning policies involves hierarchical decision structures. However, training such methods in practice may lead to poor generalization, with either sub-policies executing actions for too few time steps or devolving into a single policy altogether. In our work, we introduce an alternative approach to learn such skills \textit{sequentially} without using an overarching hierarchical policy. We propose this method in the context of environments where a major component of the objective of a learning agent is to prolong the episode for as long as possible. We refer to our proposed method as \textit{Sequential Soft Option Critic}. We demonstrate the utility of our approach on navigation and goal-based tasks in a flexible simulated 3D navigation environment that we have developed. We also show that our method outperforms prior methods such as Soft Actor-Critic and Soft Option Critic on various environments, including the Atari River Raid environment and the Gym-Duckietown self-driving car simulator.
\end{abstract}

\section{INTRODUCTION}
\noindent
Reinforcement Learning (RL) in the past decade achieved unprecedented success in multiple domains ranging from playing simple Atari games~\cite{mnih2013playing} to learning complex strategies and defeating pro players in Starcraft~\cite{vinyals2017starcraft} and Go ~\cite{silver2017mastering}. However, the problem of learning meaningful skills in reinforcement learning remains an open question. The options framework~\cite{sutton1999between} provides a method to automatically extract temporally extended skills for a long horizon task with the use of options, which are sub-policies that can be leveraged by some other policy in a hierarchical manner. The process of learning such temporal abstractions has been widely studied in the broad domain of hierarchical reinforcement learning~\cite{BerliacHierachialRL2019}. In this paper, we provide an alternate approach for learning options sequentially without a higher-level policy and show a better performance on navigation tasks. 

In the options framework, each option is defined by a tuple of policies, initiation states, and termination states. The set of termination states of an option is determined by a termination function that maps the state space to its class membership probability of the termination state set. Various advancements in the options framework, like the option critic architecture~\cite{bacon2017option}, have significantly improved the convergence of the overall algorithm, but most recent works focus on a fixed set of options that are hard to scale in practical scenarios.

\begin{figure}
    \centering
    \subfigure[]{\includegraphics[width=0.44\linewidth]{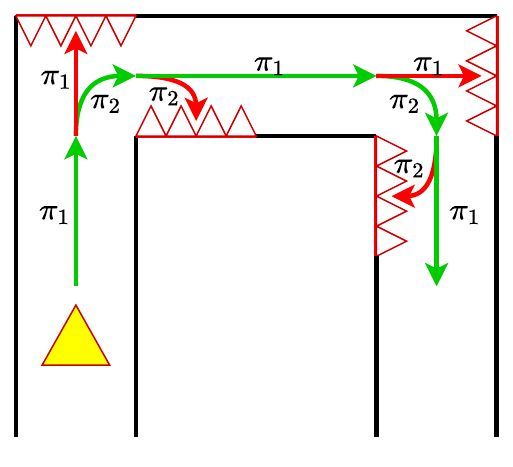}\label{fig:illustration}} 
    \subfigure[]{\includegraphics[width=0.40\linewidth, angle=270, origin=c]{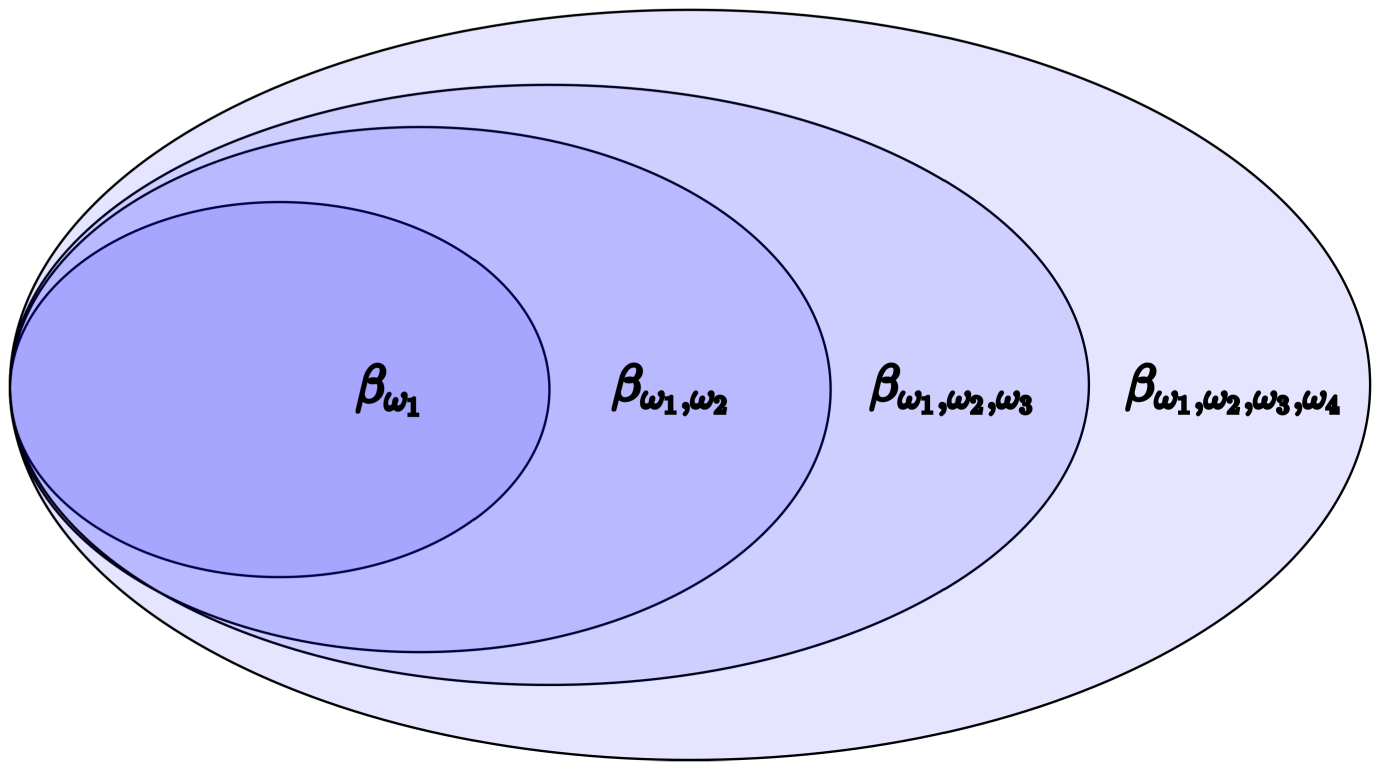}\label{fig:beta_label}}
    \caption{(a) An illustration of policies learned in our approach. Different policies learn different skills out of necessity to traverse the environment and not terminate the episode. The trajectory spawned by the sequence of policies that correctly learns to avoid terminating the episode is shown in green. The policy $\pi_1$ is used when the agent is in the middle of the corridor. When the end of the corridor is reached, $\pi_2$ is selected to take a turn and avoid a collision. (b) A representation of the state space partitioned by the termination functions of the options. Each oval corresponds to the set of states classified as non-termination states by the corresponding nested termination function.}
\end{figure}

Humans excel at learning skills because while performing tasks, they can apply an inordinate amount of prior information~\cite{dubey2018investigating}. For an RL agent to learn efficiently in complex environments, it also needs to rely on its previously learned knowledge to continually improve its overall policy. In this spirit, we propose an algorithm to learn policies sequentially so that during the training of any policy, the knowledge obtained till then, in the form of the already trained previous policies, can be leveraged to inform the learning of the current policy. We refer to our method as 
Sequential Soft Option Critic (SSOC) that is designed to operate in a framework wherein a major component of the goal of the RL agent is to learn diverse skills so as to prolong the episode and survive, as shown in Fig.~\ref{fig:illustration}. This behavior is incentivized by emitting a reward signal of $-1$ when the episode ends and a reward signal of $0$ for every other time step.

Conceptually one may draw parallels between our approach and curriculum learning~\cite{bengio2009curriculum}. The idea behind curriculum learning is to train a model with a curriculum consisting of a sequence of tasks of increasing complexity rather than simply allowing the model to learn the original task from scratch. In practice, it has been demonstrated that this approach significantly outperforms traditional learning methods~\cite{elman1993learning}. However, the major disadvantage of curriculum-based RL is that it is generally expensive to create a comprehensive curriculum, if not outright impractical. In our approach, when a new option is added, its policy is trained only in states in which the previously trained options are expected to perform poorly. This is done by using the options' termination functions to effectively partition the state space so that each option can learn an optimal policy for some subset of the state space, which is an easier task to accomplish. The options are then chained together with the termination state of the previous option serving as the initial state of the next.

We evaluate our proposed method on three environments:
\begin{enumerate*}[label=(\roman*)]
    \item a flexible 3D navigation environment developed by us,
    \item the Duckietown self-driving car simulator \cite{gym_duckietown}, and
    \item the Atari River Raid environment.
\end{enumerate*}

By conducting extensive experiments, we establish that our method outperforms the Soft Actor-Critic algorithm~\cite{haarnoja2018soft} and its options counterpart, the Soft Option Critic~\cite{lobo2019soft}. Our main contributions are as follows.

\noindent
\textbf{(1)} We propose a new approach called `Sequential Soft Option Critic' for training options in environments where a primary objective of the agent is to prolong the length of the episode.

\noindent
\textbf{(2)} We demonstrate the utility of sequentially adding new skills without a policy over options, with experimental results that outperform prior methods in the task of navigation in our simulated 3D environment and the Ducktietown and Atari River Raid environments.

\noindent
\textbf{(3)} We show that our approach can learn skills to solve complex tasks involving high-level goals in the navigation environment outperforming prior methods.

\section{RELATED WORK}
\noindent
The concept of temporal abstraction in reinforcement learning has been extensively explored in various works, from humble beginnings with options framework~\cite{sutton1999between}, feudal learning~\cite{dayan1993feudal}, hierarchical abstract machines~\cite{parr1998reinforcement}, and the MAXQ hierarchical learning algorithm~\cite{dietterich2000hierarchical}, to recent endeavors in imagination augmented agent learning with variational temporal abstraction~\cite{kim2019variational}. Approaches like feudal networks~\cite{vezhnevets2017feudal} based on feudal learning fused a manager network to choose the direction of navigation in the latent space when learning workers (sub-policies).

The option-critic architecture~\cite{bacon2017option} builds on top of the options framework and makes use of the policy over options to learn its corresponding $Q$ function. This acts as a critic and is used to update the termination functions of the options. Recently Soft Option Actor-Critic Architecture (SOAC) extended this approach by appending intrinsic rewards into the framework~\cite{li2020soac}. Unlike~,\cite{bacon2017option} which uses option critic to compute gradients for each sub-policy, in this paper, we study learning sub-policies one at a time.

Hierarchical reinforcement learning with off-policy~\cite{nachum2018data} provided a data-efficient method of training hierarchical policies.  Hindsight experience replay~\cite{andrychowicz2017hindsight} has widely been adopted for training policies in sparse reward environments and has also been recently used in multi-level hierarchical reinforcement learning algorithms ~\cite{levy2019learning}. Hierarchical reinforcement learning has also shown remarkable success in very complex domains like playing the game of Starcraft~\cite{vinyals2017starcraft}, although the sub-policies were trained separately and combined together by a master policy, the agent learned to play the game like a pro player~\cite{pang2019reinforcement}. Meta-learning approaches~\cite{frans2018meta} focused on training a meta controller, which would be frequently re-initialized such that it can learn to control the trained sub-policies.

Our approach amounts to partitioning the state space based upon how well a policy performs in it, much like previous iterative approaches~\cite{mankowitz2016iterative}. Our method also shares some commonalities with the deep skill chaining algorithm~\cite{bagaria2019option} on how new options are added to the existing set of options. Deep skill chaining sequentially learns local skills by chaining them backward from a goal state. However, in our approach, skills are learned to complement the previously acquired ones such that the agent can traverse to various unseen states. This method shares a striking resemblance with curriculum-based learning approaches \cite{bengio2009curriculum}, but here the curriculum naturally arises from the necessity of traversing the environment. The use of nested termination functions in our framework is inspired by the continual learning architecture in progressive neural networks~\cite{rusu2016progressive}. We make use of the Soft Actor-Critic algorithm (SAC)~\cite{haarnoja2018soft} that is based on a maximum entropy reinforcement learning framework~\cite{ziebart2010modeling} for training all our policy networks.

\section{BACKGROUND}
\noindent
A Markov decision process (MDP) is defined by the tuple $(\mathcal{S}, \mathcal{A}, p, r, \gamma)$, where $\mathcal{S}$ is the state space, $\mathcal{A}$ is the action space, $p$ is the state transition probability $p(s_{t+1}|s_t, a_t)$ of going to the next state $s_{t+1}\in\mathcal{S}$ from state $s_t\in\mathcal{S}$ given an action $a_t\in\mathcal{A}$, $r$ is the reward function $r(s_t, a_t) \in \mathbb{R} $ that provides a reward signal as the agent traverses the environment, and $\gamma\in{[0, 1)}$ is the discount factor. The aim is to find an optimal policy $\pi^*$ such that
\begin{equation}
\pi^*= \argmax_\pi \sum_{t=0}^T \mathsf{E}_{(s_t, a_t)\sim\rho_\pi}[\gamma^tr(s_t, a_t)] \:, \label{eq:1}
\end{equation}
where $\rho_\pi$ is the state-action distribution induced by a policy $\pi$. For every policy $\pi$, one can define its corresponding $Q$ value function:
\begin{equation}
    \label{eq:Q_value}
    Q_\pi(s_t, a_t) = r(s_t, a_t) + \gamma \mathsf{E}_{s_{t+1} \sim p(\cdot|s_t, a_t)} V_\pi(s_{t+1})\:,
\end{equation}
where $V_\pi$ is the value function defined by
\begin{equation}
    \label{eq:V_value}
    V_\pi(s_t) = \mathsf{E}_{a_t\sim \pi}[Q_\pi(s_t, a_t)]\:.
\end{equation}

\subsection{Soft Actor-Critic}
\noindent
The soft actor-critic algorithm~\cite{haarnoja2018soft} is an off-policy entropy-based reinforcement learning algorithm. The main idea of the entropy-based learning approach is to maximize the entropy of the policy along with the reward. A straightforward way of doing this is by making the reward function depend on the current policy's entropy. Making the reward proportional to the entropy incentivizes greater exploration of the environment, ensuring the policy is far less likely to get stuck in a local optimum. Thus, the optimal policy is defined as
\begin{equation}
    \pi^* = \argmax_\pi \sum_{t=0}^T \mathsf{E}_{(s_t, a_t)\sim\rho_\pi}\gamma^t [r(s_t, a_t) + \alpha\mathcal{H}(\cdot|s_t)]\:,  \label{eq:3}
\end{equation}
where $\alpha$ is the temperature variable that accounts for the importance of the entropy and $\mathcal{H}(\cdot|s_t)$ is the entropy of the policy. The above formulation reduces to the standard reinforcement learning objective as $\alpha \rightarrow{0}$. It is shown that by iteratively using the soft policy evaluation and soft policy improvement, the policy convergences to the optimal policy $\pi^*$ \cite{haarnoja2018soft}.

The temperature variable $\alpha$ can be treated as a trainable parameter for better performance of the algorithm \cite{haarnoja2018soft_var}. This gives the algorithm the flexibility to dictate the relative importance of the policy's entropy. As $\alpha$ decreases, the policy becomes more deterministic in nature. We make use of this property of the soft actor-critic to decide when new policies should be added.

\subsection{The Options Framework}
\noindent
The idea of temporally extended actions has been introduced by \cite{sutton1999between}. An option $\omega \in \Omega$ is defined by the tuple $(\mathcal{I}_\omega, \pi_\omega, \beta_\omega)$, where $\Omega$ is the set of options, $\pi_\omega$ is the policy corresponding to the option, $\mathcal{I}_\omega (\subseteq\mathcal{S})$ is the set of states where the option can be initialized, and $\beta_\omega: S \rightarrow [0, 1]$ is the termination function of the option. In a typical options framework, $k$ sub-policies, $\pi_{\omega_1}, \pi_{\omega_2},...\pi_{\omega_k}$ are initialized, each with its corresponding $\mathcal{I_\omega}$ and $\mathcal{\beta_\omega}$, along with a policy over options $\pi_\Omega$. In the \textit{call-and-return} approach $\pi_\Omega$ chooses an option $\omega$ and $\pi_\omega$ executes actions till it terminates with probability $\beta_\omega(s_t)$ for a given state $s_t$ and the control then returns back to $\pi_\Omega$. The state transition dynamics are given by
\begin{align}
\label{eq:option}
& P(s_{t+1}, \omega_{t+1}| s_t, \omega_t)  \nonumber \\ & =   \sum_a \pi_{\omega_t}(a|s_t)P(s_{t+1}|s_t, a) \big[ (1  - \beta_{\omega_t}(s_{t+1})) 1_{\omega_t = \omega_{t+1}} \nonumber \\
& \qquad \qquad \qquad \quad \quad \qquad + \beta_{\omega_t}(s_{t+1}) \pi_\Omega(\omega_{t+1}|s_{t+1}) \big] \:.
\end{align}
Unlike other approaches where the next option is chosen by a master policy $\pi_\Omega$, in our proposed approach, the next option to be executed depends on the termination state space of the previous options.

\section{THE PROPOSED ALGORITHM}

\subsection{Class of environments}
\noindent
In this paper, we consider environments in which a major component of the task of the agent is to prolong the duration of the episode, which is incentivized by using a simple reward function $r(s_t, a_t, s_{t+1})$, whose value is $-1$ if $s_{t+1}$ is the final state of the episode, and $0$ otherwise. Many real-life problems can be effectively cast as reinforcement learning problems with such a reward function and minimal information from the environment. Examples of such applications include autonomous vehicle navigation while avoiding a collision \cite{kahn2018self, shalev2016safe} and drone navigation \cite{kang2019generalization}. Tasks that require the agent to maintain an equilibrium in an ever-changing environment may be cast into our framework using this simple reward function. Practical examples may involve tasks like assembly line automation with increasing levels of complexity.

A consequence of such a reward signal is that as the agent fails less frequently in the process of learning better policies, it becomes harder to train it owing to the increasing sparsity of the failure states. Our approach is designed to overcome this by learning new policies near states where failure occurs without disturbing the already learned policies.

In our proposed approach, the policies learned need only be locally optimal. The main challenge of the learning algorithm then becomes determining how likely it is for the current policy to fail so as to switch to another policy. Unlike other option learning algorithms, the proposed strategy is to sequentially train one option at a time. To learn from a minimal amount of data while training, we do not train a (master) policy over the options but rely on the termination functions of individual options to determine which option should be chosen for execution. The states in which a new policy is trained are determined by the termination functions, which are treated as binary classifiers. New policies are learned only in states that are classified as termination states for all the previous options, and this is achieved by nesting the termination functions of the options. We explain this procedure in detail in the later subsections.

Ideally, one can continue to add more options into the set of all options until $V_\Omega(s_t) \approx 0$, where state $s_t$ belongs to the marginal state distribution induced by the set of options $\Omega$ learned by the algorithm. However, from a pragmatic perspective, we threshold the maximum number of options to be learned. We now describe in detail how each component of the framework is adopted in our approach.

\subsection{Training Policies}
\noindent
In the proposed approach, the policies are trained sequentially, i.e., firstly, a single policy and its termination function are trained until semi-optimality, only then another policy is added along with its termination function and trained until semi-optimality, and so on. Once a policy is trained, it is not modified again. Later in this section, we go into detail about how each policy is trained and what we consider to be the semi-optimality of a policy.

Corresponding to any given policy, depending on its performance, the state space is partitioned into termination and non-termination states for that policy. Suppose we have a trained option $\omega_1$. The next option $\omega_2$ need only be trained in those states that are classified as termination states by $\beta_{\omega_1}$, i.e., the new policy is only focused on learning to operate in states where the previous policy failed. In order to make use of the previously learned policy, we also incentivize the new policy to traverse towards the states that are deemed non-termination states by $\beta_{\omega_1}(s_t)$. For this, in general, given trained options $\omega_1, \ldots, \omega_{i-1}$, we train a new policy $\pi_{\omega_i}$ with the following reward function
\begin{equation}
    r_{\pi_{\omega_i}}(s_t, a_t, s_{t+1}) =
    \begin{cases}
    1 \quad \text{ if } \widetilde{\beta}_{... \omega_{i-1}}(s_{t+1},.) = 0,\\
    r(s_t, a_t, s_{t+1}) \quad \text{ otherwise,}
    \end{cases}
    \label{eq:plus_one_reward}
\end{equation}
where $\widetilde{\beta}_{... \omega_{i-1}}$ is a `nested termination function', which acts as a termination function corresponding to all the previously trained options $\omega_1, \ldots, \omega_{i-1}$ together. The learning of this nested termination function is described in detail in the next subsection. The reason for giving a $+1$ `inter-option' reward whenever the new policy enters states with $\widetilde{\beta}_{... \omega_{i-1}}(s_{t+1},.) = 0$ is because the desired objective is to incentivize the new policy to enter states where it can easily switch to some other policy that has already been fully trained and hence, is presumably more capable of traversing those states. This encourages the agent to leverage previously gained knowledge instead of trying to relearn it. The training of the new policy is limited to those state-action pairs that are vital for prolonging the episode and cannot be delegated to previous policies. It is to be noted that if the policy remains in states for which $\widetilde{\beta}_{... \omega_{i-1}}(s_{t+1},.) = 0$, i.e, termination states of previous policies, then the reward given to the policy is the unchanged $\{0,-1\}$ sparse reward of the environment.

This modification of the reward function described by Equation \ref{eq:plus_one_reward} is applied for training every policy other than the first one. We included ablation studies in our experiments to show that this inter-option reward is an important reason behind our approach outperforming the Soft Actor-Critic algorithm in the navigation environment.

Each policy is updated as in soft-actor critic, by using the information projection on the exponential of the soft Q-value
\begin{equation}
    \pi^{new}_\omega = \argmin_{\pi'\in \prod} D_{KL}\left(\pi'(.|s_t) \Big\|
    \frac{\exp(\frac{1}{\alpha}Q^{\pi_{\omega}}(s_t, .))}{Z^{\pi_{\omega}}(s_t)}\right),
\end{equation}
where $Z^{\pi_\omega}(s_t)$ normalizes the distribution. 

In this work, we make use of the trainable $\alpha$ of the soft actor-critic algorithm as a measure to decide whether to stop training a policy and add a new option. $\alpha$ is updated so as to minimize the following cost function~\cite{haarnoja2018soft_var}
\begin{equation}
    \mathit{J}(\alpha) = \E_{a_t \sim \pi_t}[-\alpha \log \pi_t (a_t|s_t) - \alpha \bar{\mathcal{H}}],
\end{equation}
where $\bar{\mathcal{H}}$ is a hyperparameter set to $\,-$dim$(\mathcal{A})$, where $\mathcal{A}$ is the action space. If $\alpha$ is constrained to satisfy $\alpha \geq 0$, the optimal value of $\alpha$ that satisfies the above objective is $0$, since the entropy $\E_{a_t \sim \pi_t} [-\log \pi_t (a_t|s_t)] \geq 0 > \bar{\mathcal{H}}$. So the value of $\alpha$ monotonically decreases as the training progresses.

As the value of $\alpha$ decreases, there is a smaller incentive for the agent to maximize the entropy. This results in the learned policy becoming less exploratory and more deterministic. We determine a policy to be sufficiently trained for it to function as a semi-optimal policy when $\alpha < \alpha_{min}$, where $\alpha_{min}\in[0, 1]$ is a threshold. All new sub-policies are initialized with $\alpha=1$ to ensure maximum exploration near states where policies are initialized. Essentially we augment a new policy when the temperature variable $\alpha$ of the currently trained policy is low enough to warrant it to be considered an optimal policy near its initialization states.

Once a policy $\pi_\omega$ is trained enough to be deemed semi-optimal, we fix that policy and no longer train it. The rationale for this is that it is much easier to train new options in the context of fixed already trained options rather than trying to learn new options while simultaneously updating old options.

\begin{figure*}
    \centering
    \vspace{0.2cm}
    \subfigure[]{\includegraphics[width=0.27\linewidth]{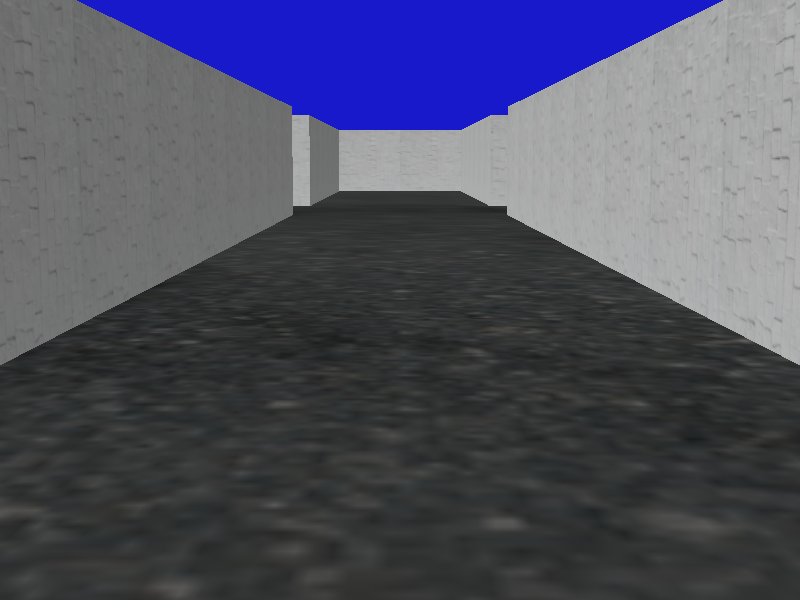}} 
    \subfigure[]{\includegraphics[width=0.27\linewidth]{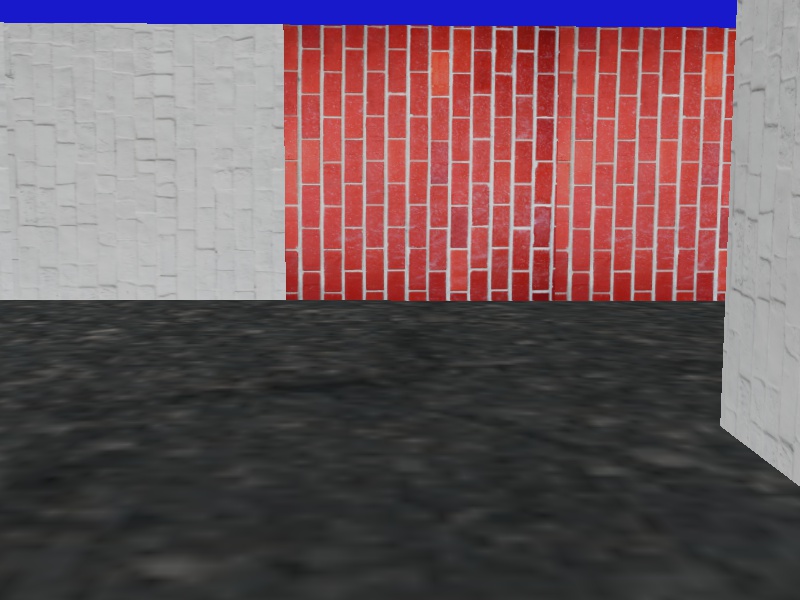}\label{envpic_color}}
    \subfigure[]{\includegraphics[width=0.27\linewidth]{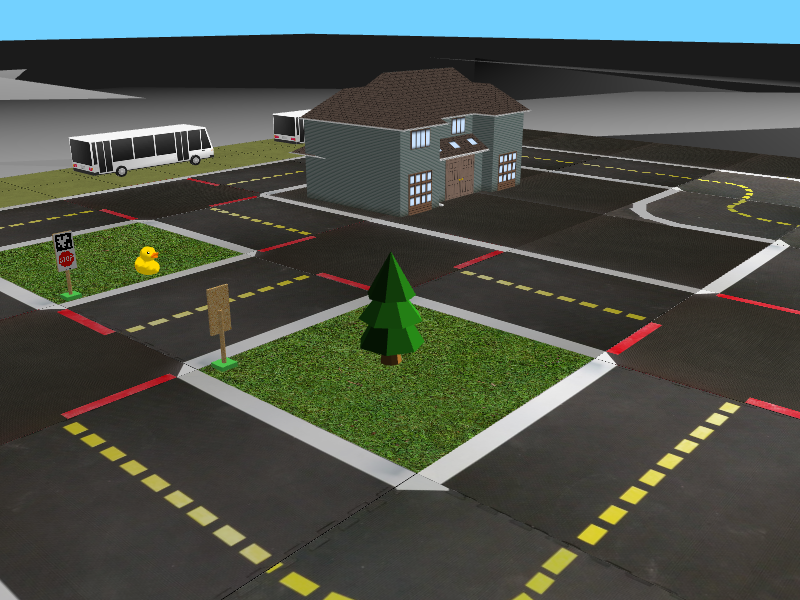}}
    \caption{(a) \& (b) The input to the neural network in our 3D navigation environment. These images obtained from the simulation are scaled (and also transformed to grayscale in the case of (a)) and appended with $K$ previous images, and sent to the agent as input. (c) A view of the Duckietown environment.}
    \label{fig:minipage2}
\end{figure*}

\subsection{Learning Termination Functions}
\noindent
For each option $\omega$, $\, \widetilde{\beta}_\omega(s_t)$ is either $1$ or $0$, depending on whether option $\omega$ should terminate at state $s_t$ or not. This can be implemented by learning a real-valued function $\beta_\omega (s_t)$ and using it as a binary classifier. Suppose the learning algorithm has already trained $i-1$ options, and the aim is to learn a new option $\omega_i$. We train its new policy as if a single policy is being trained in all states $s_t$ satisfying $\widetilde{\beta}_\omega(s_t) = 1$ for all $\omega \in \Omega_{old} = \{ \omega_1, \omega_2, ... \omega_{i-1} \}$, i.e., in states that are classified as termination states by all the previous options. For this, rather than training $\beta_{\omega_i}$ for the $i^{th}$ option, we train a nested termination function $\beta_{\omega_1, \omega_2,..., \omega_i}$, which is the termination classifier for the set of options $\{ \omega_1, \omega_2, ..., \omega_i \}$. As the new policies are trained in the termination states of the previous set of options, the set of non-termination states for the set of options keeps expanding, as shown in Fig. \ref{fig:beta_label}. Without nested functions, as the number of options increases, it becomes more difficult to accurately update the individual termination functions of each policy, which in turn makes it difficult to incorporate new skills. In the proposed approach, as long as the last nested termination function is correctly updated, new policies can always be learned in the termination states of that function. Using $\widetilde{\beta}_{\omega_1 ..., \omega_i}(s_t)$ and $\widetilde{\beta}_{\omega_1 ..., \omega_{i-1}}(s_t)$, we can obtain $\widetilde{\beta}_{\omega_i}(s_t)$ as
\begin{equation}
    \widetilde{\beta}_{\omega_i}(s_t) = (1-\widetilde{\beta}_{\omega_1 ..., \omega_{i-1}}(s_t)) \vee \widetilde{\beta}_{\omega_1 ..., \omega_i}(s_t)\enspace. \label{eq:beta}
\end{equation}
The above equation imposes a simple constraint on choosing options for execution by not allowing a new policy to execute actions in states which are classified as a non-terminating state by the previous nested termination function since that means there is already some trained option more suited to execute in that state.

Since $\tilde{\beta}_{\omega_1, \dots, \omega_i}$ has to take values in $0$ or $1$ (non-termination and termination respectively), we learn corresponding continuous functions $\beta_{\omega_1, \dots, \omega_i}$ with range $[0,1]$ and use a threshold to assign $\tilde{\beta}_{\omega_1, \dots, \omega_i}$ the value $1$ if $\beta_{\omega_1, \dots, \omega_i}$ exceeds the threshold, $0$ otherwise.

$\beta_{\omega_{1} \dots \omega_{i}}$ is trained like a standard $Q$-value function using the negative of the reward signal emitted by the environment,
\begin{equation}
    r_{\beta}(s_t, a_t, s_{t+1}) =
    \begin{cases}
    1 & \parbox[t]{.55\linewidth}{if $s_{t+1}$ is the last state in the episode, and}\\
    0 & \text{otherwise.}\\
    \end{cases} \label{eq:6}
\end{equation}
We overload the notation of $\beta_{\omega_{1} \dots \omega_i}$ by using $\beta_{\omega_{1} \dots \omega_i}(s_t, a_t)$ as the termination function rather than $\beta_{\omega_{1} \dots \omega_i}(s_t)$, since we train it like a Q-value function and because it is a better estimator. This function is learned using the update rule:
\begin{equation}
    \beta_{\omega_{1 ...}}(s_t, a_t) \leftarrow r_{\beta}(s_t, a_t, s_{t+1}) + \gamma_{\beta}\beta_{\omega_{1 ...}}(s_{t+1}, a_{t+1}), \label{eq:td_update}
\end{equation}
where $\gamma_\beta \in [0, 1]$ is the termination discount factor. It largely depends on the problem domain and directly impacts the newer policies that are trained by influencing the partition of the termination and non-termination states.

As new policies are incorporated, $r_\beta(s_t, a_t)$ becomes an increasingly sparse reward since the agent learns strategies to avoid failure, and episodes become relatively longer. In such instances, simply training $\beta_{\omega_{1} \dots \omega_i}(s_t)$ using the above update can bias it to predict all states as non-termination states and thus effectively prevent new policies from learning. To avoid this, after the new policy has been semi-optimally trained, the termination function is trained using the binary cross-entropy loss as
\begin{align}
\beta_{\omega_{1 ...}} \leftarrow \argmin_{\beta_{\omega_{1} \dots \omega_i}} & - \E_{(s_t, a_t)} \, [ y_t \log \beta_{\omega_{1} \dots \omega_i}(s_t, a_t) \nonumber \\
& +\; (1-y_t) \log (1-\beta_{\omega_{1} \dots \omega_i}(s_t, a_t) ) ] \:.
\label{eq:bce_loss}
\end{align}
Here the labels $y_t$ are the solution to (\ref{eq:td_update}) if the policy that generates the actions is fixed. They are obtained by unrolling a trajectory $\{s_0, ..., s_{T}\}$ and labeling each state $s_t$ with $y_t = \gamma_\beta^{T-t}$, where $\gamma_\beta$ is the discount factor.

As we fix the policies after training them, the cross-entropy update gives us a better estimation as compared to the previous temporal difference updates. To make $\beta_{\omega_{1} \dots \omega_i}(s_t, a_t)$ unbiased, an equal number of probable termination and non-termination states are sampled for training. The major advantage of adopting nested termination functions is that we only need to modify the latest termination function to correctly reflect if a state is a termination state or not. That will then be used to determine the states in which the next new policy will be trained. Given a training set $\Omega$ of options, an option is chosen during execution according to the constraint given in (\ref{eq:beta}).

\begin{figure*}
\centering
\vspace{0.3cm}
\subfigure[3D environment with 3 options]{\includegraphics[width=0.29\textwidth]{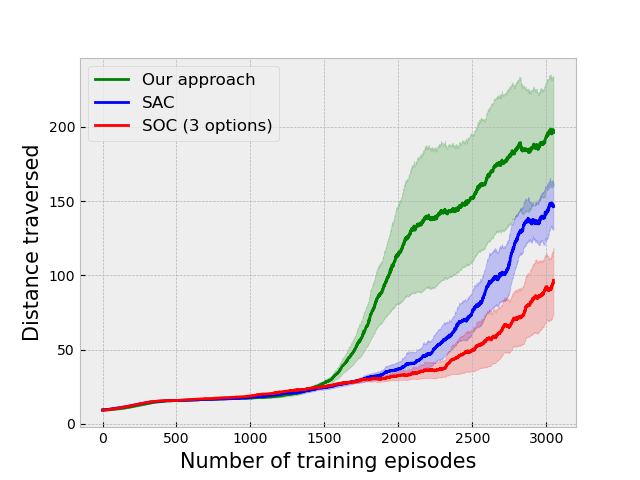}\label{fig:navigation_3}} \quad
\subfigure[Color 3D environment with 3 options]{\includegraphics[width=0.29\textwidth]{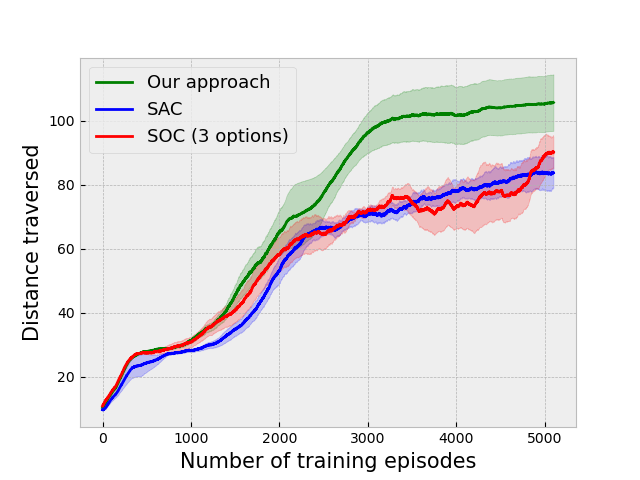}\label{fig:navigation_4}} \quad 
\subfigure[Duckietown environment with 2 options]{\includegraphics[width=0.29\textwidth]{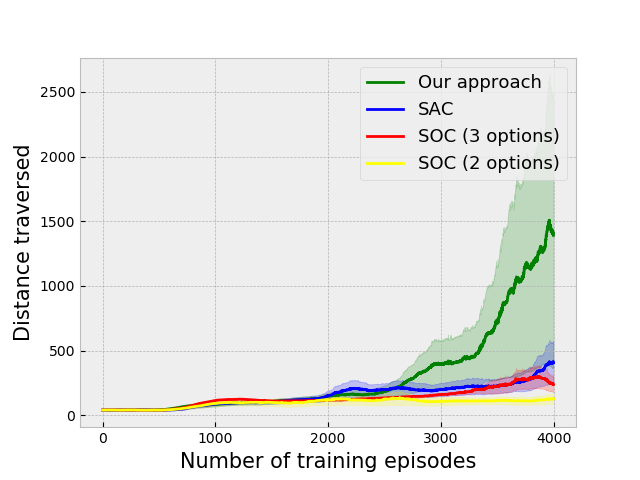}\label{fig:duckie_results}} \\
\subfigure[3D navigation environment with 4 options]{\includegraphics[width=0.3\textwidth]{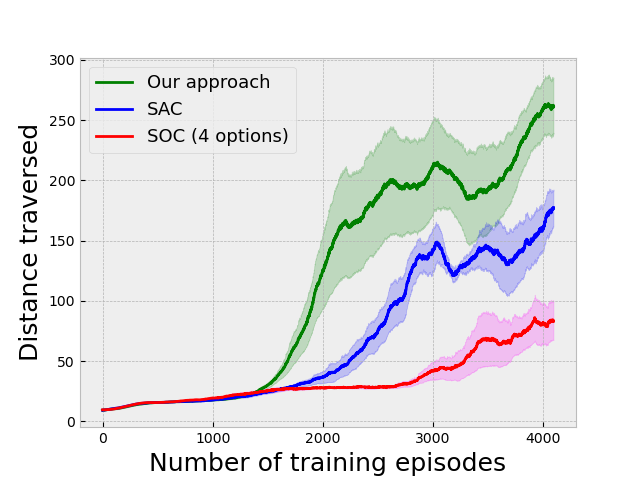}\label{fig:policy_4}} \quad
\subfigure[Atari environment with 2 options]{\includegraphics[width=0.29\textwidth]{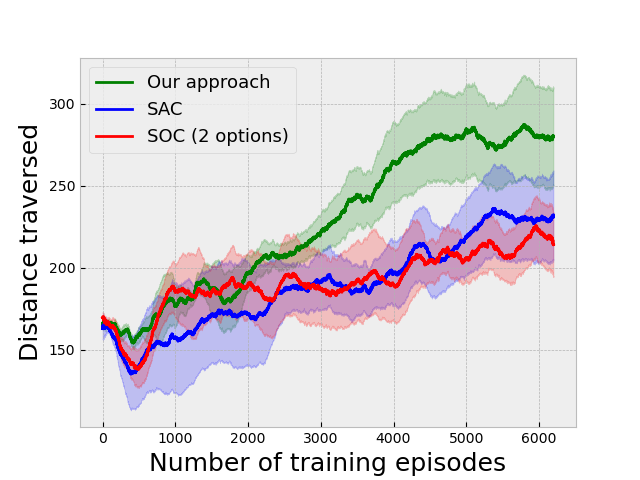}\label{fig:atari_soc}} \quad
\subfigure[Atari River Raid environment with 3 options]{\includegraphics[width=0.29\textwidth]{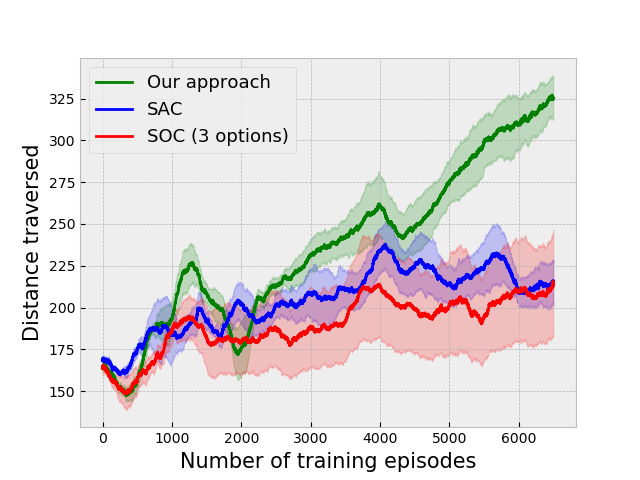}\label{fig:river_raid_3}} \\
\subfigure[Goal based navigation in the 3D environment]{\includegraphics[width=0.29\textwidth]{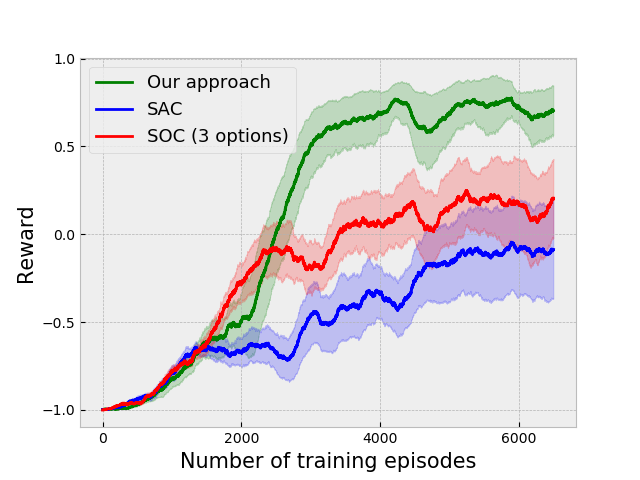}\label{fig:arrow_env}} \quad
\subfigure[Duckie Town environment without inter-option rewards]{\includegraphics[width=0.29\textwidth]{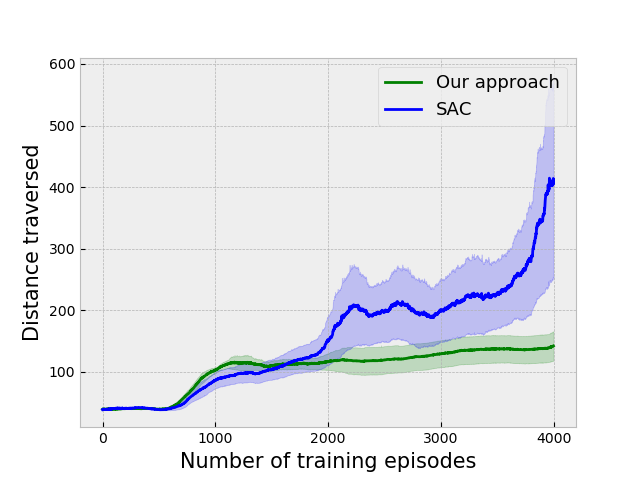}\label{fig:Duckie_ablation}} \quad 
\subfigure[Instruction based navigation in the 3D environment]{\includegraphics[width=0.29\textwidth]{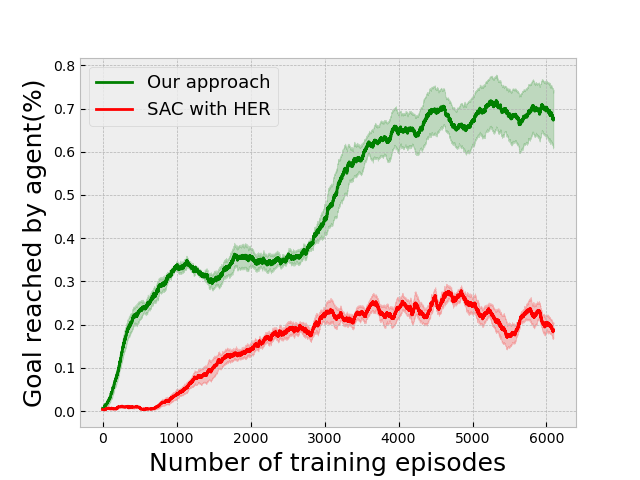}\label{fig:goal_navigation_plot}}
\caption{The results obtained on various environments}
\end{figure*}

\section{EXPERIMENTAL RESULTS}
\noindent
\begin{figure}
  \centering
  \includegraphics[width=0.6\linewidth]{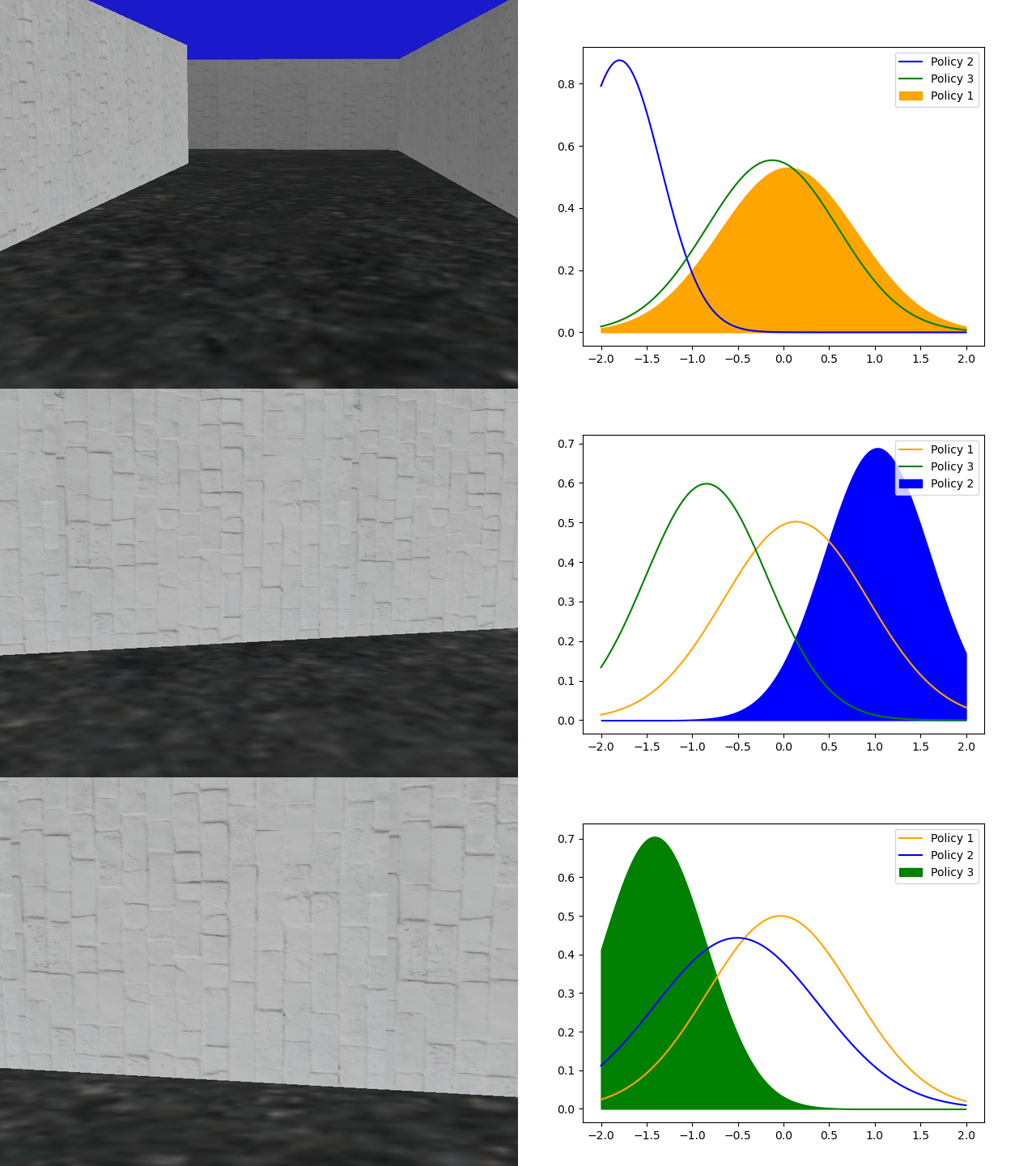}
  \caption{Outputs of the three learned policies when fed the corresponding input shown on the left. $\pi_{\omega_1}$: orange, $\pi_{\omega_2}$: blue, $\pi_{\omega_3}$: green. Each policy outputs a Gaussian distribution, the active policy is the one filled with color. A positive value for the output action corresponds to turning right, and a negative value indicates a left turn.}
  \label{fig:learnt_policies}
\end{figure}
We compare our approach on several environments against the Soft Actor critic (SAC) \cite{haarnoja2018soft} and the Soft Option-Critic (SOC) \cite{lobo2019soft} algorithms. Each experimental plot has been plotted by taking the mean of 5 different experiments, and the corresponding bounds are given $\pm \text{std} /2$.

\subsection{The 3D navigation environment}
\noindent
We have created a 3D simulated environment using the Panda3D game engine \cite{goslin2004panda3d}. This collision avoidance environment consists of long corridors twisting and turning as the agent navigates inside as a vehicle that moves with a constant velocity. The primary challenge is that agent has to understand whether the left or right turn is coming up as taking a wrong turn will inevitably result in a collision and end the episode. The results on this environment are given in Fig.~\ref{fig:navigation_3}. In Fig. \ref{fig:learnt_policies} we show a snapshot of the learned policies navigating in the environment. The first option simply learns to move straight in the environment, while the second and third policies learn to take the correct turns.

We have also validated our approach on a goal-based version of this environment, in which the agent is also given as input a number representing a destination and is guided towards it by arrows placed in the environment. The results for this setting are given in Fig. \ref{fig:arrow_env}. We have not used an inter-option reward for this setting. We have also validated our method on an instruction-based task in which the agent is given a one-hot vector denoting the direction to take at an intersection, with the results given in Fig. \ref{fig:goal_navigation_plot} compared against Soft Actor-Critic with Hindsight Experience Replay. For this setting, we have also used a $+1$ reward for going in the correct direction and $-\frac{1}{2}$ for going in the wrong direction. Additionally, Fig. \ref{fig:navigation_4} shows the performance of our approach on a colored version of the environment in which the agent has to take the correct turn depending on the color of the walls in front of it. It should understand the concept of taking a turn in the direction of the colored wall if the color is green and in the opposite direction if the color is red.

\subsection{Other environments}
\noindent
We have tested our algorithm on two more environments: (i) Gym-Duckietown \cite{gym_duckietown}, a self-driving car simulator that is a complex environment consisting of multiple immediate turns and various objects like houses, trees, etc., resulting in a large variance in the observations from the environment, and (ii) the Atari River Raid environment, a top-down shooting game in which the goal is to maneuver a plane to destroy or avoid obstacles. The results of the experiments on these environments are given in Fig. \ref{fig:duckie_results} and \ref{fig:atari_soc} respectively.

Fig. 
\ref{fig:Duckie_ablation} show the results of our algorithm when the inter-option reward is removed, showing that it is an essential component of our algorithm in a navigation environment.

\addtolength{\textheight}{-1cm}

\subsection{Implementation}
\noindent
The policy and Q functions are implemented as neural networks that take, at each time step, scaled images seen by the agent from the past $K$ time steps, where $K=10$ for the navigation environment and $K=4$ for the Duckietown and Riverraid environments. In environments where the color of the image is not explicitly required as part of the goal specification, the input images are transformed to grayscale before being passed to the networks. The outputs of the policy networks are the parameters of a Gaussian distribution for the case of continuous actions in the navigation and Duckietown environments, and parameters of a categorical distribution for the case of discrete actions in the Riverraid environment.

Two critic networks are used, and the smaller of their predicted values is taken as the $Q$ value to tackle overestimation. Additionally, two additional networks are used as target (soft) $Q$ networks, whose parameters follow those of the critics as exponential moving averages with smoothing coefficient $\tau$.

\begin{table}
    \vspace{0.3cm}
    \caption{Hyperparameters}
    \label{tab:hyperparams}
    \centering
    \begin{tabular}{ll} 
      \hline
      \textbf{Parameter} & \textbf{Value}\\
      \hline
      learning rate & $3 \times 10^{-4}$\\
      discount factor($\gamma$) & $0.99$\\
      replay buffer size & $10^4$\\
      target smoothing coefficient($\tau$) & $0.005$\\
      number of frames in input state ($K$) & $10$ \\
      batch size & $16$ \\
      alpha threshold ($\alpha_{min}$) & $0.1$ \\
      termination discount factor ($\gamma_\beta$) & $0.95$\\
      \hline
    \end{tabular}
\end{table}

\section{CONCLUSIONS}
\noindent
In this paper, we proposed an algorithm called `Sequential Soft Option Critic' that allows adding new skills dynamically without the need for a higher-level master policy. This can be applicable to environments where a primary component of the objective is to prolong the episode. We show that this algorithm can be used to effectively incorporate diverse skills into an overall skill set, and it outperforms prior methods in several environments.

\section{ACKNOWLEDGMENTS}
Authors would like to thank Shubham Gupta for many useful discussions on this topic.





\bibliographystyle{./IEEEtran}
\bibliography{IEEEabrv,bibfile}

\begin{thebibliography}{10}
\providecommand{\url}[1]{#1}
\csname url@rmstyle\endcsname
\providecommand{\newblock}{\relax}
\providecommand{\bibinfo}[2]{#2}
\providecommand\BIBentrySTDinterwordspacing{\spaceskip=0pt\relax}
\providecommand\BIBentryALTinterwordstretchfactor{4}
\providecommand\BIBentryALTinterwordspacing{\spaceskip=\fontdimen2\font plus
\BIBentryALTinterwordstretchfactor\fontdimen3\font minus
  \fontdimen4\font\relax}
\providecommand\BIBforeignlanguage[2]{{%
\expandafter\ifx\csname l@#1\endcsname\relax
\typeout{** WARNING: IEEEtran.bst: No hyphenation pattern has been}%
\typeout{** loaded for the language `#1'. Using the pattern for}%
\typeout{** the default language instead.}%
\else
\language=\csname l@#1\endcsname
\fi
#2}}

\bibitem{mnih2013playing}
V.~Mnih, K.~Kavukcuoglu, D.~Silver, A.~G.~I. Antonoglou, D.~Wierstra, and
  M.~Riedmiller, ``Playing atari with deep reinforcement learning,''
  \emph{Neural Information Processing Systems}, 2013.

\bibitem{vinyals2017starcraft}
O.~Vinyals, T.~Ewalds, S.~Bartunov, P.~Georgiev, A.~S. Vezhnevets, M.~Yeo,
  A.~Makhzani, H.~K{\"u}ttler, J.~Agapiou, J.~Schrittwieser, \emph{et~al.},
  ``Starcraft ii: A new challenge for reinforcement learning,'' \emph{arXiv
  preprint arXiv:1708.04782}, 2017.

\bibitem{silver2017mastering}
D.~Silver, J.~Schrittwieser, K.~Simonyan, I.~Antonoglou, A.~Huang, A.~Guez,
  T.~Hubert, L.~Baker, M.~Lai, A.~Bolton, \emph{et~al.}, ``Mastering the game
  of go without human knowledge,'' \emph{Nature}, vol. 550, no. 7676, pp.
  354--359, 2017.

\bibitem{sutton1999between}
R.~S. Sutton, D.~Precup, and S.~Singh, ``Between mdps and semi-mdps: A
  framework for temporal abstraction in reinforcement learning,''
  \emph{Artificial intelligence}, vol. 112, no. 1-2, pp. 181--211, 1999.

\bibitem{BerliacHierachialRL2019}
Y.~Flet-Berliac, ``The promise of hierarchical reinforcement learning,''
  \emph{The Gradient}, 2019.

\bibitem{bacon2017option}
P.-L. Bacon, J.~Harb, and D.~Precup, ``The option-critic architecture,'' in
  \emph{Thirty-First AAAI Conference on Artificial Intelligence}, 2017.

\bibitem{dubey2018investigating}
R.~Dubey, P.~Agrawal, D.~Pathak, T.~Griffiths, and A.~Efros, ``Investigating
  human priors for playing video games,'' in \emph{International Conference on
  Machine Learning}, 2018, pp. 1349--1357.

\bibitem{bengio2009curriculum}
Y.~Bengio, J.~Louradour, R.~Collobert, and J.~Weston, ``Curriculum learning,''
  in \emph{Proceedings of the 26th annual International Conference on Machine
  Learning}, 2009, pp. 41--48.

\bibitem{elman1993learning}
J.~L. Elman, ``Learning and development in neural networks: The importance of
  starting small,'' \emph{Cognition}, vol.~48, no.~1, pp. 71--99, 1993.

\bibitem{gym_duckietown}
M.~Chevalier-Boisvert, F.~Golemo, Y.~Cao, B.~Mehta, and L.~Paull, ``Duckietown
  environments for openai gym,''
  \url{https://github.com/duckietown/gym-duckietown}, 2018.

\bibitem{haarnoja2018soft}
T.~Haarnoja, A.~Zhou, P.~Abbeel, and S.~Levine, ``Soft actor-critic: Off-policy
  maximum entropy deep reinforcement learning with a stochastic actor,'' in
  \emph{International Conference on Machine Learning}, 2018, pp. 1861--1870.

\bibitem{lobo2019soft}
E.~Lobo and S.~Jordan, ``Soft options critic,'' \emph{arXiv preprint
  arXiv:1905.11222}, 2019.

\bibitem{dayan1993feudal}
P.~Dayan and G.~E. Hinton, ``Feudal reinforcement learning,'' in \emph{Advances
  in neural information processing systems}, 1993, pp. 271--278.

\bibitem{parr1998reinforcement}
R.~Parr and S.~J. Russell, ``Reinforcement learning with hierarchies of
  machines,'' in \emph{Advances in neural information processing systems},
  1998, pp. 1043--1049.

\bibitem{dietterich2000hierarchical}
T.~G. Dietterich, ``Hierarchical reinforcement learning with the maxq value
  function decomposition,'' \emph{Journal of artificial intelligence research},
  vol.~13, pp. 227--303, 2000.

\bibitem{kim2019variational}
T.~Kim, S.~Ahn, and Y.~Bengio, ``Variational temporal abstraction,'' in
  \emph{Advances in Neural Information Processing Systems}, 2019, pp.
  11\,570--11\,579.

\bibitem{vezhnevets2017feudal}
A.~S. Vezhnevets, S.~Osindero, T.~Schaul, N.~Heess, M.~Jaderberg, D.~Silver,
  and K.~Kavukcuoglu, ``Feudal networks for hierarchical reinforcement
  learning,'' in \emph{Proceedings of the 34th International Conference on
  Machine Learning-Volume 70}.\hskip 1em plus 0.5em minus 0.4em\relax JMLR.
  org, 2017, pp. 3540--3549.

\bibitem{li2020soac}
C.~Li, X.~Ma, C.~Zhang, J.~Yang, L.~Xia, and Q.~Zhao, ``Soac: The soft option
  actor-critic architecture,'' \emph{arXiv preprint arXiv:2006.14363}, 2020.

\bibitem{nachum2018data}
O.~Nachum, S.~S. Gu, H.~Lee, and S.~Levine, ``Data-efficient hierarchical
  reinforcement learning,'' in \emph{Advances in Neural Information Processing
  Systems}, 2018, pp. 3303--3313.

\bibitem{andrychowicz2017hindsight}
M.~Andrychowicz, F.~Wolski, A.~Ray, J.~Schneider, R.~Fong, P.~Welinder,
  B.~McGrew, J.~Tobin, O.~P. Abbeel, and W.~Zaremba, ``Hindsight experience
  replay,'' in \emph{Advances in neural information processing systems}, 2017,
  pp. 5048--5058.

\bibitem{levy2019learning}
A.~Levy, G.~Konidaris, R.~Platt, and K.~Saenko, ``Learning multi-level
  hierarchies with hindsight,'' in \emph{Proceedings of International
  Conference on Learning Representations}, 2019.

\bibitem{pang2019reinforcement}
Z.-J. Pang, R.-Z. Liu, Z.-Y. Meng, Y.~Zhang, Y.~Yu, and T.~Lu, ``On
  reinforcement learning for full-length game of starcraft,'' in
  \emph{Proceedings of the AAAI Conference on Artificial Intelligence},
  vol.~33, 2019, pp. 4691--4698.

\bibitem{frans2018meta}
K.~Frans, J.~Ho, X.~Chen, P.~Abbeel, and J.~Schulman, ``Meta learning shared
  hierarchies,'' in \emph{International Conference on Learning
  Representations}, 2018.

\bibitem{mankowitz2016iterative}
D.~J. Mankowitz, T.~A. Mann, and S.~Mannor, ``Iterative hierarchical
  optimization for misspecified problems (ihomp),'' \emph{arXiv preprint
  arXiv:1602.03348}, 2016.

\bibitem{bagaria2019option}
A.~Bagaria and G.~Konidaris, ``Option discovery using deep skill chaining,'' in
  \emph{the NeurIPS 2019 Workshop on Deep Reinforcement Learning}, 2019.

\bibitem{rusu2016progressive}
A.~A. Rusu, N.~C. Rabinowitz, G.~Desjardins, H.~Soyer, J.~Kirkpatrick,
  K.~Kavukcuoglu, R.~Pascanu, and R.~Hadsell, ``Progressive neural networks,''
  \emph{arXiv preprint arXiv:1606.04671}, 2016.

\bibitem{ziebart2010modeling}
B.~D. Ziebart, ``Modeling purposeful adaptive behavior with the principle of
  maximum causal entropy,'' 2010, aAI3438449.

\bibitem{haarnoja2018soft_var}
T.~Haarnoja, A.~Zhou, K.~Hartikainen, G.~Tucker, S.~Ha, J.~Tan, V.~Kumar,
  H.~Zhu, A.~Gupta, P.~Abbeel, and S.~Levine, ``Soft actor-critic algorithms
  and applications,'' in \emph{arXiv preprint arXiv:1812.05905}, 2018.

\bibitem{kahn2018self}
G.~Kahn, A.~Villaflor, B.~Ding, P.~Abbeel, and S.~Levine, ``Self-supervised
  deep reinforcement learning with generalized computation graphs for robot
  navigation,'' in \emph{2018 IEEE International Conference on Robotics and
  Automation (ICRA)}.\hskip 1em plus 0.5em minus 0.4em\relax IEEE, 2018, pp.
  1--8.

\bibitem{shalev2016safe}
S.~Shalev-Shwartz, S.~Shammah, and A.~Shashua, ``Safe, multi-agent,
  reinforcement learning for autonomous driving,'' \emph{arXiv preprint
  arXiv:1610.03295}, 2016.

\bibitem{kang2019generalization}
K.~Kang, S.~Belkhale, G.~Kahn, P.~Abbeel, and S.~Levine, ``Generalization
  through simulation: Integrating simulated and real data into deep
  reinforcement learning for vision-based autonomous flight,'' in \emph{2019
  International Conference on Robotics and Automation (ICRA)}.\hskip 1em plus
  0.5em minus 0.4em\relax IEEE, 2019, pp. 6008--6014.

\bibitem{goslin2004panda3d}
M.~Goslin and M.~R. Mine, ``The panda3d graphics engine,'' \emph{Computer},
  vol.~37, no.~10, pp. 112--114, 2004.

\bibitem{harb2018waiting}
J.~Harb, P.-L. Bacon, M.~Klissarov, and D.~Precup, ``When waiting is not an
  option: Learning options with a deliberation cost,'' in \emph{Thirty-Second
  AAAI Conference on Artificial Intelligence}, 2018.

\bibitem{schaul2015prioritized}
T.~Schaul, J.~Quan, I.~Antonoglou, and D.~Silver, ``Prioritized experience
  replay,'' \emph{arXiv preprint arXiv:1511.05952}, 2015.

\bibitem{christodoulou2019soft}
P.~Christodoulou, ``Soft actor-critic for discrete action settings,''
  \emph{arXiv preprint arXiv:1910.07207}, 2019.

\end{thebibliography}


\clearpage
\appendix[LEARNING SKILLS TO NAVIGATE WITHOUT A MASTER]
\subsection{Pseudocode}

Algorithm \ref{alg:train} contains the pseudocode for training a new option given a set of already trained options. Algorithm \ref{alg:eval} describes how to use the options to traverse the environment.

\begin{algorithm}[H]
\caption{Training an option}
\label{alg:train}
\begin{algorithmic}
\STATE {\bfseries Input:} $\Omega = \{\omega_1, \omega_2, ..., \omega_{k-1}\}$, a set of $k-1$ options, and threshold $\alpha_{min}$
\STATE Initialize environment $\mathcal{E}$
\STATE Initialize replay buffer $\mathcal{R}$
\STATE Initialize new option $\omega_k = \{\pi_{\omega_k}, \beta_{\omega_1, ...\omega_k} \}$
\STATE $\alpha_{\omega_k} \leftarrow 1$
\WHILE{$\alpha_{\omega_k} \geq \alpha_{min}$}
    \STATE $\widetilde{\beta}_{t} \leftarrow 1$
    \IF{$\Omega \neq \emptyset$}
        \STATE Get $a_t$ from $\Omega$ for state $s_t$ and update $i$ as shown in Algorithm 2
        \STATE $\widetilde{\beta_{t}} \leftarrow $ classify using $\beta_{\omega_1,..,\omega_{k-1}}(s_t, a_t)$
    \ENDIF
    \IF{$\widetilde{\beta}_{t} = 1$}
        \STATE $i \leftarrow 1$
        \STATE $a_t \sim \pi_{\omega_k}(s_t)$
    \ENDIF
    \STATE Execute action $a_t$ in the environment $\mathcal{E}$ to get the next state $s_{t+1}$ and the reward $r_t$
    \STATE $r_{\beta_t} \leftarrow -r_t$
    \IF{$\Omega \neq \emptyset$ and $\widetilde{\beta}_t = 1$}
        \STATE Get $a_{t+1}$ from $\Omega$ for state $s_{t+1}$ as shown in \\ Algorithm 2
        \STATE $\widetilde{\beta}_{t+1} \leftarrow $ classify using $\beta_{\omega_1,..,\omega_{k-1}}(s_{t+1}, a_{t+1})$\\
        \IF{$\widetilde{\beta}_{t+1} = 0$}
            \STATE $r_t \leftarrow 1$
        \ENDIF
    \ENDIF
    \IF{$\widetilde{\beta}_t = 0$}
        \STATE Store transition tuple $(s_t, a_t, s_{t+1}, r_t, r_{\beta_t})$ in replay buffer $\mathcal{R}$
        \STATE Update $\pi_{\omega_k}$ and $\alpha_{\omega_k}$ with $(s_t, a_t, s_{t+1}, r_t)$ sampled from $\mathcal{R}$ with the soft actor critic update
        \STATE Update $\beta_{\omega_1, ...\omega_k}$ with $(s_t, a_t, s_{t+1}, r_{\beta_t})$ sampled from $\mathcal{R}$ with TD update
    \ENDIF
\ENDWHILE

\STATE Add $\omega_k$ to the set of options $\Omega$
\STATE Initialize buffer $\mathcal{D}$
\STATE Roll-out trajectories in the environment as described in Algorithm 2 and store them in $\mathcal{D}$
\STATE Train $\beta_{\omega_1, ...\omega_k}$ with $(s_t, a_t, y_t)$ sampled from $\mathcal{D}$ using the binary cross entropy loss as shown in equation (\ref{eq:bce_loss})
\end{algorithmic}
\end{algorithm}

\hspace{30mm}

\begin{algorithm}[H]
\caption{Executing trained options}
\label{alg:eval}
\begin{algorithmic}
    \STATE {\bfseries Input:} $\Omega = \{\omega_1, \omega_2, ..., \omega_k\}$ a set of $k$ options
    \WHILE{not last time step}
        \STATE $a_t \sim \pi_{\omega_i}(s_t)$
        \STATE $\widetilde{\beta}_t \leftarrow $ classify using $\beta_{\omega_1, ..., \omega_k}(s_t, a_t)$
        \STATE $i \leftarrow k$
        \IF{$\widetilde{\beta_t} = 0$}
            \WHILE{$i>1$}
                \STATE $a_{i-1} \sim \pi_{\omega_{i-1}}(s_t)$
                \STATE $\widetilde{\beta}_{i-1} \leftarrow $ classify using $\beta_{..., \omega_{i-1}}(s_t, a_{i-1})$
                \IF{$\beta_{i-1} = 1$}
                    \STATE break
                \ENDIF
                \STATE $a_t \leftarrow a_{i-1}$
                \STATE $\widetilde{\beta}_t \leftarrow \widetilde{\beta}_{i-1}$
                \STATE $i \leftarrow i-1$
            \ENDWHILE
        \ENDIF
        \STATE Execute action $a_t$ in the environment $\mathcal{E}$ to get the next state $s_{t+1}$
        \STATE $s_t \leftarrow s_{t+1}$
    \ENDWHILE
\end{algorithmic}
\end{algorithm}

\subsection{The developed 3D environment}
\label{sec:3d_env}

\begin{figure}[H]
    \centering
    \includegraphics[width=0.49\textwidth]{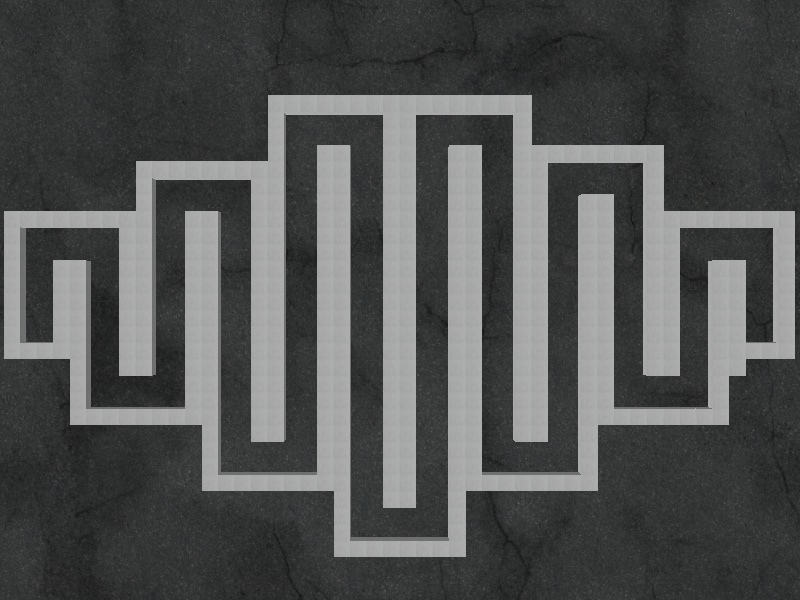}
    \caption{A top down view of the navigation environment. The agent is initialized at one of the two corners randomly at the start of each episode.}
    \label{fig:my_label}
\end{figure}

\begin{figure*}
    \centering
    \subfigure[]{\includegraphics[width=0.45\textwidth]{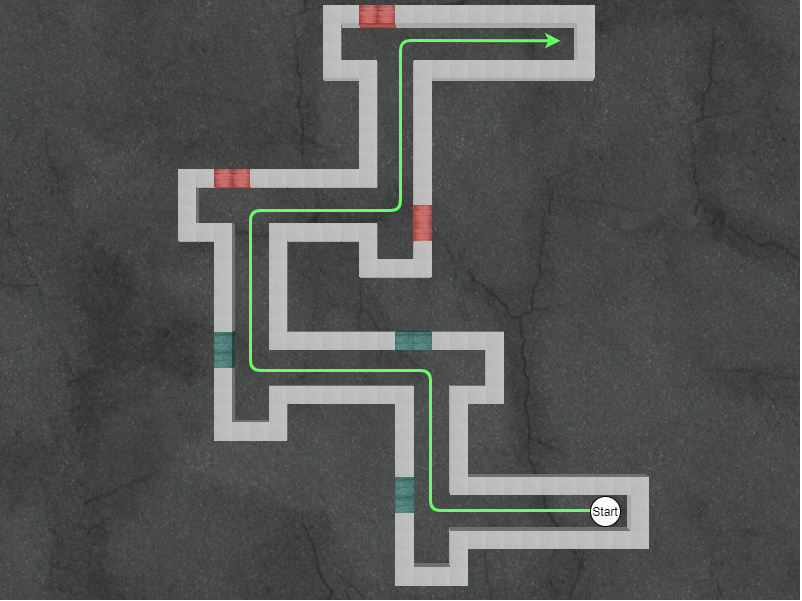} \label{fig:color_maze}} \quad
    \subfigure[]{\includegraphics[width=0.45 \textwidth]{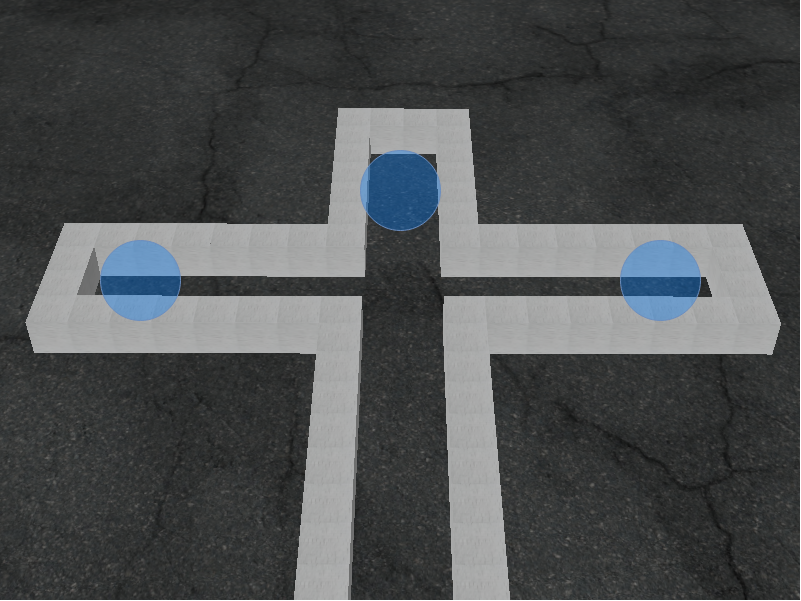} \label{fig:goal_based_camera}}
    \subfigure[]{\includegraphics[width=0.45\textwidth]{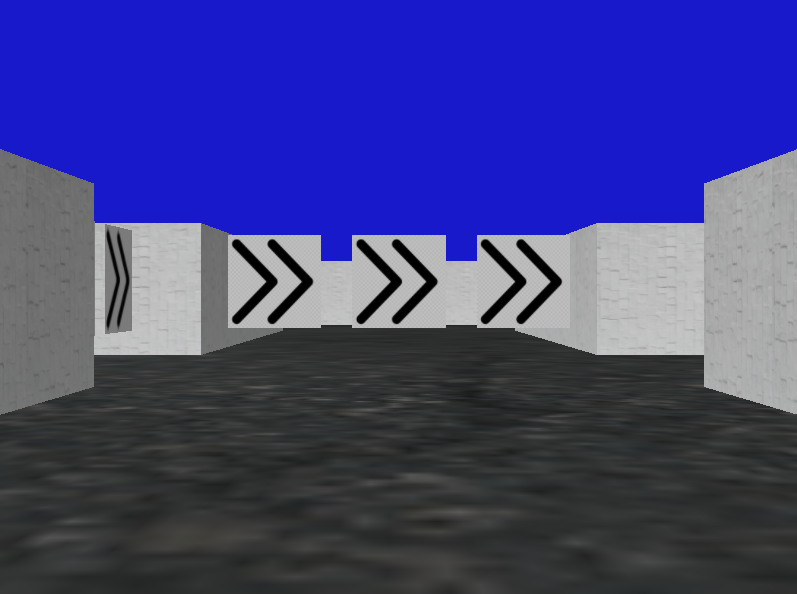} \label{fig:3D_Instr}} \quad
    \subfigure[]{\includegraphics[width=0.45\textwidth]{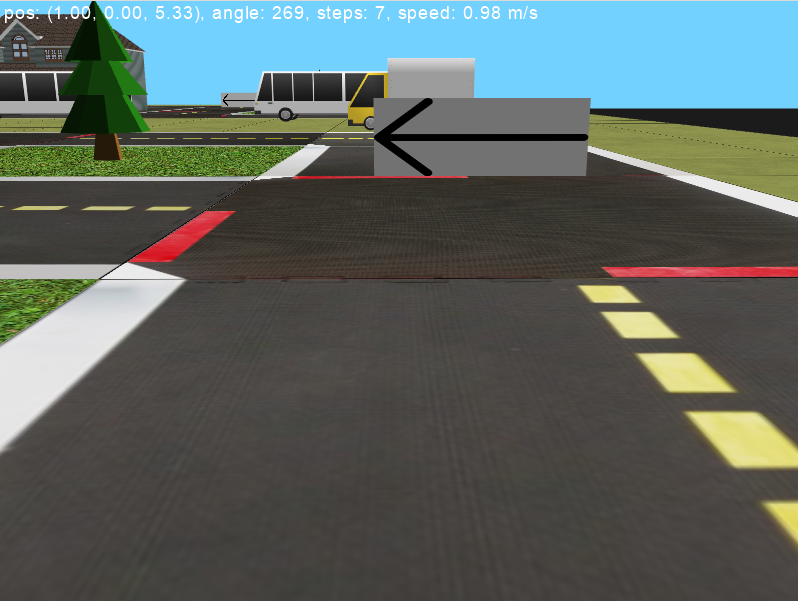} \label{fig:Duckie_Instr}} 
    \subfigure[]{\includegraphics[width=0.45\textwidth]{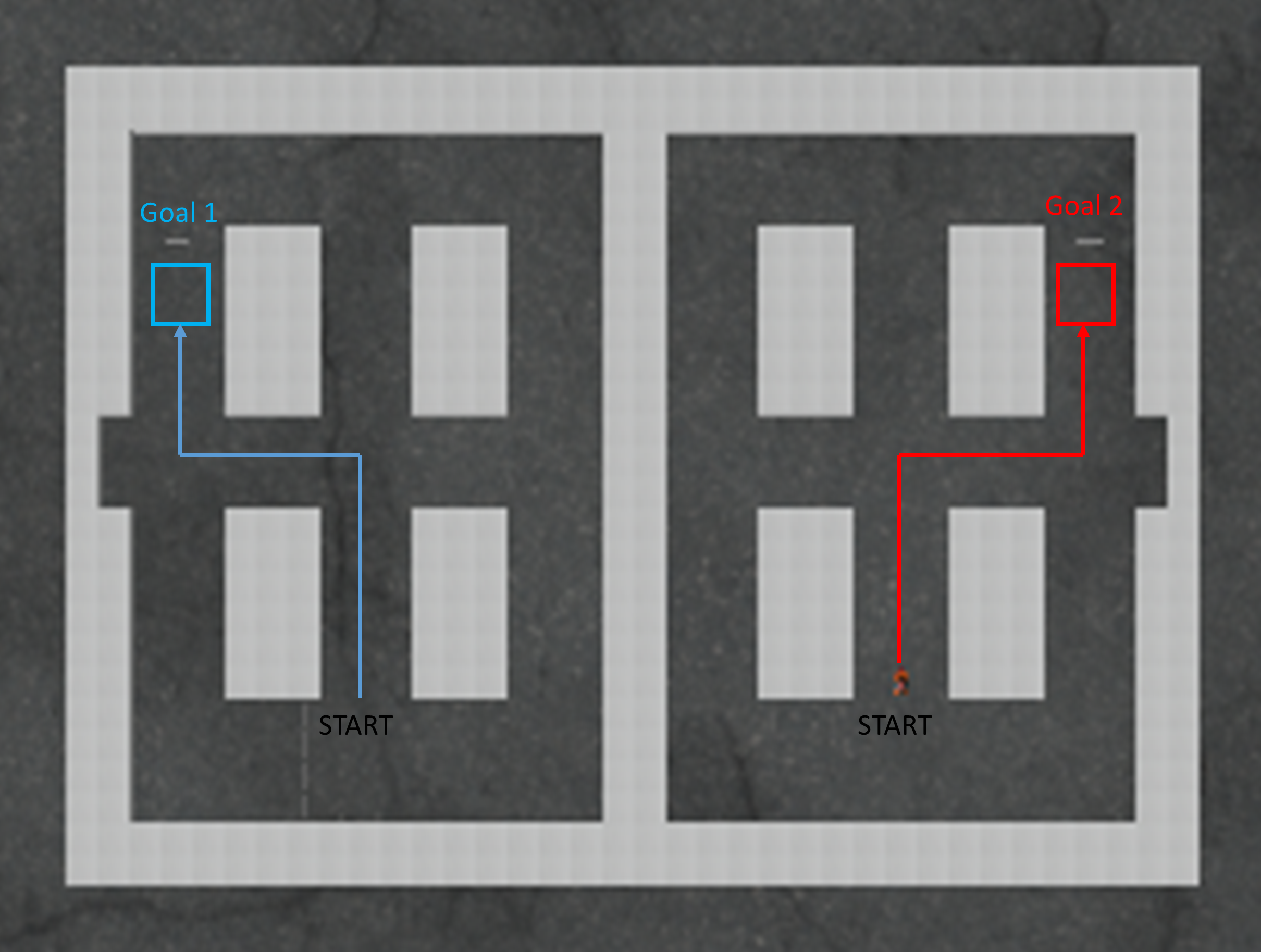}\label{fig:3D_topview}} \qquad \:
    \subfigure[]{\includegraphics[width=0.42\textwidth]{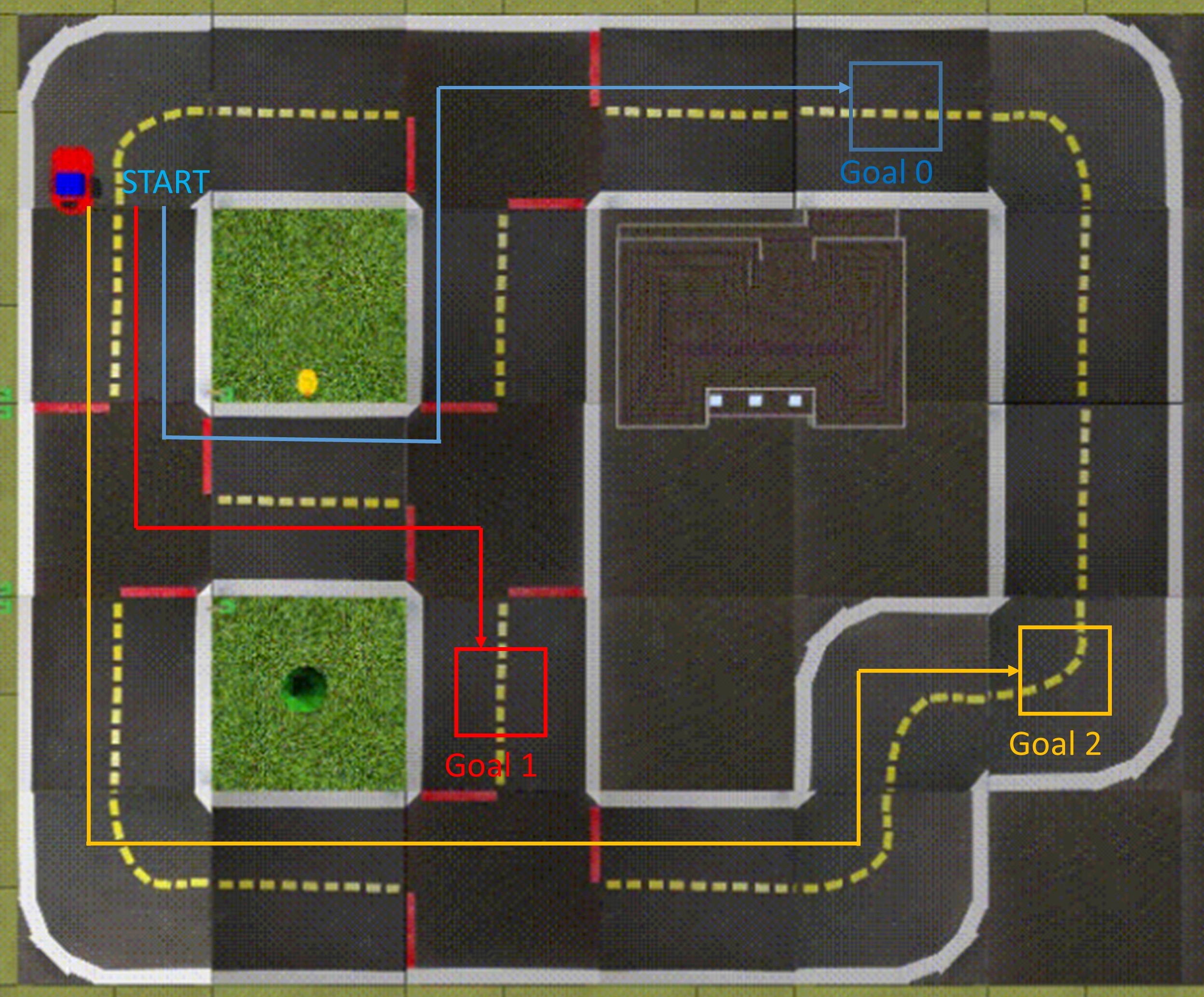} \label{fig:Duckie_top}}
    \caption{(a) A top down view of the color navigation environment. The green arrow indicates the correct path that should be taken by the vehicle. (b) A birds' eye view of the goal based 3D navigation environment in which the agent takes a one-hot vector as an instruction. The blue circles represent the corresponding goal state. (c) \& (d): Screenshots of the 3D environment and Duckietown environment, with the arrows pointing towards the direction in which the agent has to go. (e) \& (f) The view of these environments from above. The coloured lines show the paths the agent is supposed to take for different goals.}
\end{figure*}

For testing our algorithm, we have created a custom simulated 3D navigation environment using the Panda3D game engine \cite{goslin2004panda3d}. Here we present the details of this environment.

In the developed 3D navigation environment, the agent is initialized at either the extreme left or the extreme right corridor randomly. The goal of the agent is to make its way to the other end (as shown in Figure~\ref{fig:my_label}) without colliding with the walls.

The agent receives the past $K$ gray-scale images re-scaled to $36\times64$ pixels as input and outputs an action $a_t \in \mathbb{R}$, every time a collision occurs and the episode is reset. The agent takes an action every $0.25$ seconds in the environment and receives a continuous stream of $0$ rewards in each time-step except when it collides with the walls, in which case the episode ends with a $-1$ reward. Overall, the environment is a rather simplistic one with mono-coloured white walls, minimalist texture and ambient lighting.

An interesting aspect of this environment is that the single source lighting creates different shades on the walls depending on the direction they are facing, so the agent has to learn to turn depending on its understanding of what turn is coming up while disregarding the shade of the wall.

In the coloured navigation environment (Figure~\ref{fig:color_maze}) the agent has to follow a path specified by the colour of the walls in front of it to take the correct direction. Two contiguous block segments are coloured to indicate the direction that the vehicle is supposed to take and the color determines if the direction is correct or not. Green indicates that the turn is the correct one while red is incorrect and the vehicle should take the turn in the opposite direction.

\subsection{Other environments}
Other than our 3D navigation environment, we have validated our approach on two environments: (i) Gym-Duckietown, and (ii) Atari River Raid environment.

Gym-Duckietown is a self-driving car simulator that is a complex environment consisting of multiple immediate turns, making navigation challenging. It also includes various objects like houses, trees, etc. These visuals result in a large variance in the observations from the environment. For our experiments, we modify the reward function to emit $-1$ whenever the agent steers off the road and $0$ during every other time step when the agent is on the road.

River Raid is a top-down shooting game where the goal is to maneuver a plane to destroy or avoid obstacles like tankers, helicopters, jets, and bridges to continuously keep moving forward, all the while not crashing into the walls on the sides or the obstacles themselves. Destroying the obstacles rewards the player with an increase in overall points obtained in the game; the plane also needs to refill its fuel, which it loses continually as the game progresses and is refilled by hovering over fuel tanks (which can also be destroyed for obtaining points) which frequently spawn in the environment. The plane crashes when it either collides with an obstacle or runs out of fuel.

For our experiments, we modify the reward function of the environment to emit a $-1$ reward when the agent crashes and a $0$ reward for every other time step. We also restrict the agent to use a reduced number of actions while navigating in the environment such that it is not allowed to slow down the plane, and our modified action set has $6$ discrete actions the agent can take as compared to $18$ in the original gym implementation. We follow the same preprocessing of the input image and network architecture as used by \cite{mnih2013playing}. Since the reward is extremely sparse in such a setting, we store only the last $25$ time steps in the replay buffer for better training.

\begin{figure*}
    \centering
    \subfigure[3D navigation environment without internal rewards]{\includegraphics[width=0.45\textwidth]{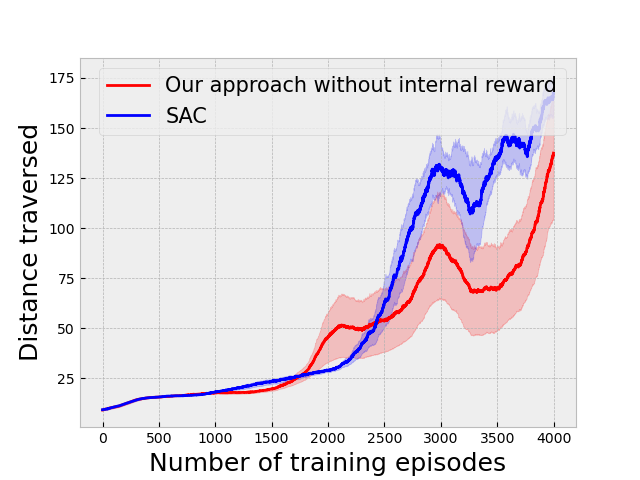}\label{fig:ablations}}
    \subfigure[Duckie Town environment with arrow instruction.]{\includegraphics[width=0.41\textwidth]{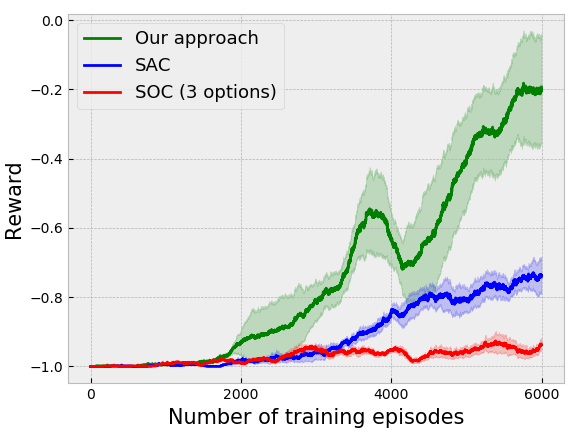} \label{fig:Duckie_Instr_Result}}
    \caption{.}
\end{figure*}

\subsection{Training agents on goal-based tasks}
 The class of environments that we described in this work captures many practical scenarios. For example, many goal-based navigation environments can be modelled using this approach. Here, we consider one such environment, shown in Figure \ref{fig:goal_based_camera}, where the objective of the agent is to navigate corridors according to a specified instruction. The instruction is given as a one-hot vector input that corresponds to the corridor the agent is supposed to navigate to (left, center or right for instructions $[1, 0, 0]$, $[0, 1, 0]$ or $[0, 0, 1]$ respectively). For this, a reward function can be defined as,
 
\begin{equation}
    r(s_t, a_t, s_{t+1}) =
    \begin{cases}
    0 & \begin{aligned}[t]
        \text{if } s_{t+1} \text{ is not last state}\\
    \end{aligned} \\
    1 & \text{if } s_{t+1} \in \mathcal{S}_{\mathrm{correct\_goal}}\\
    -1  & \text{if } s_{t+1} \in \mathcal{S}_{\mathrm{collision}}\\
    -\frac{1}{2} & s_{t+1} \in \mathcal{S}_{\mathrm{wrong\_goal}}.
    \end{cases} \label{eq:goal_based}
\end{equation}

Here, $\mathcal{S}_{\mathrm{correct\_goal}}$ is the set of states corresponding to the agent being in the correct corridor according to the given instruction, and $\mathcal{S}_{\mathrm{collision}}$ is the set of all states where the agent collides with the walls in the environment. The agent incurs a $-0.5$ reward when it enters the wrong corridor, $\mathcal{S}_{\mathrm{wrong\_goal}}$. This is given in order to deter the agent from choosing the wrong corridor, although not as great of a deterrent as colliding with the walls. However, using this reward function may also produce policies that easily fall into a local optimum of choosing a single direction by completely disregarding the instruction.

One can easily modify the proposed approach for training on this environment by including the state where the agent has taken a wrong direction in the termination state set and the state where it has taken the correct direction in the non-termination state set, for the purposes of training the termination function. The options are also made to execute for some minimum time steps $t_{min}$ before switching to another option. This is done to prevent options from switching prematurely~\cite{harb2018waiting}. We compare the results of our approach with 2 options, in Figure \ref{fig:goal_navigation_plot}, against Hindsight Experience Replay \cite{andrychowicz2017hindsight} using Soft Actor Critic with priority buffers \cite{schaul2015prioritized}. We use the same hyper-parameters as in the normal 3D navigation environment without any instructions, with the exception of $K=20$ and $\gamma_\beta=0.95$, along with $t_{min}=16$. The complete set of hyperparameters is given in Section \ref{sec:network_arch}.

Apart from the above discussed modification, and the color environment described in \ref{sec:3d_env}, we also considered another variant of a goal-based task in the 3D environment. We modified the environment by providing arrows indicating correct turn towards goal state at each intersection, as shown in Figure \ref{fig:3D_Instr}. As shown in Figure \ref{fig:3D_topview}, we considered two goals, and the blue and red line represent the correct path towards each of these goals. Here, the goal state is defined as a window of size $1\times1$ around the location shown by the colored squares. A downward arrow has been placed in the environment to indicate this goal state. The reward structure used is same as in equation \ref{eq:goal_based}. For this setting, we do not add the +1 inter-option reward when control is passed to previously learnt options. The results of our approach for this task have been shown in Figure \ref{fig:goal_navigation_plot}.

We have also considered such a goal based navigation task on a modified version of the Duckietown environment \cite{gym_duckietown}, in which we considered three goals, as shown in Figure \ref{fig:Duckie_top}. The blue, red and yellow colored lines represent the correct path for reaching goals 0, 1 and 2 respectively. The goal states are in a window of size $0.4\times0.4$ across the location represented by the square shown in the figure. The complexity of the environment and the number of goals makes this task relatively more challenging. The results for this task are shown in Figure \ref{fig:Duckie_Instr_Result}. 

\subsection{Ablations}

In our algorithm, we make use of internal rewards when training subsequent policies in a sequential manner. Each policy is incentivized to traverse to states where the previous set of policies determines that it is capable of traversing it, using the reward function:
\begin{equation}
    r_{\pi_{\omega_i}}(s_t, a_t, s_{t+1}) =
    \begin{cases}
    1 \quad \quad \quad \quad \text{if} \; \widetilde{\beta}_{... \omega_{i-1}}(s_{t+1}, .) = 0,\\
    r(s_t, a_t, s_{t+1}) \quad \quad \quad \quad \text{otherwise.}\\
    \end{cases} \label{eq:11}
\end{equation}

This greatly improves the convergence of the algorithm in scenarios where the previous policies need to be used repeatedly and rather frequently. In our 3D navigation environment, in the absence of the $+1$ reward, the new options require much more training for them to traverse to desired states by taking correct turns. We demonstrate this by training our method without providing the extra $+1$ reward using 4 options. This also shows that later options relying on already learnt knowledge in the form of previous options does in fact play a role in the performance of our algorithm. Figure \ref{fig:ablations} shows the result of our approach in the absence of the internal reward in the 3D navigation environment. Fig \ref{fig:Duckie_ablation} shows the results for the same setting in the Duckietown environment.

\subsection{Details on the Network Architecture}
\label{sec:network_arch}

\begin{figure*}
\centering
\subfigure[Image Encoder]{\includegraphics[width=0.9\textwidth]{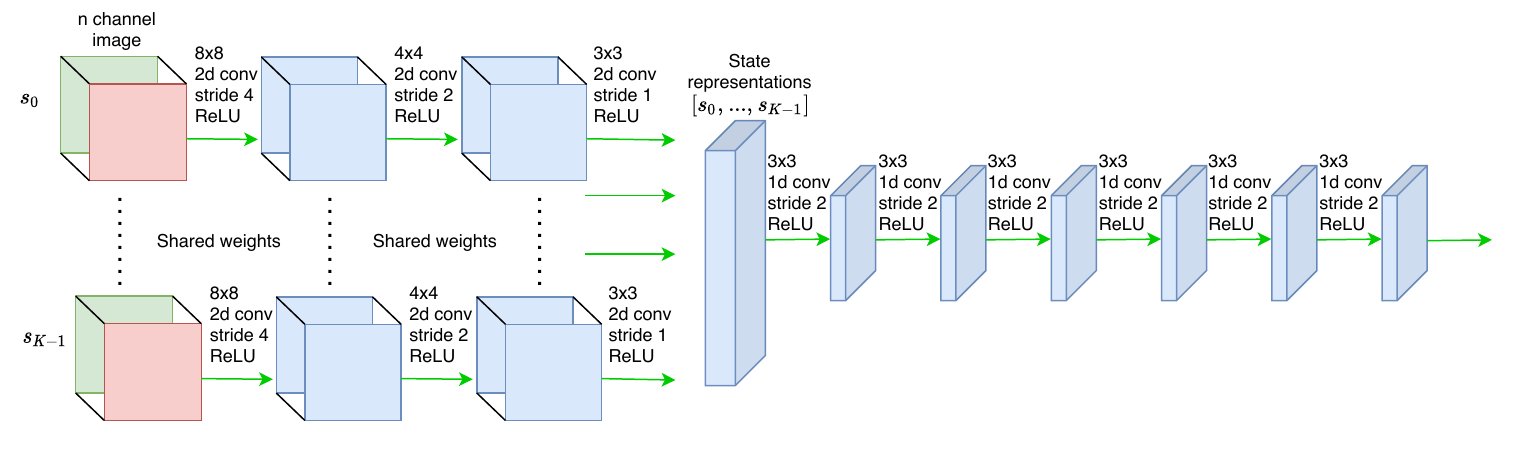}\label{fig:image_encoder}}
\subfigure[Policy Network]{\includegraphics[width=0.45\textwidth]{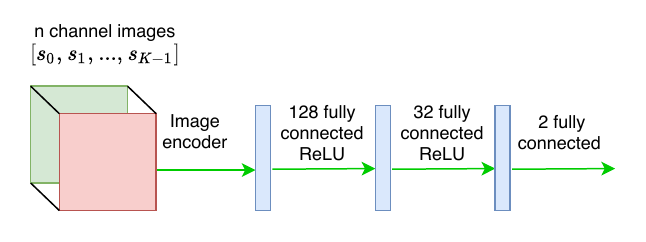}\label{fig:policy_network}}
\subfigure[Q network]{\includegraphics[width=0.45\textwidth]{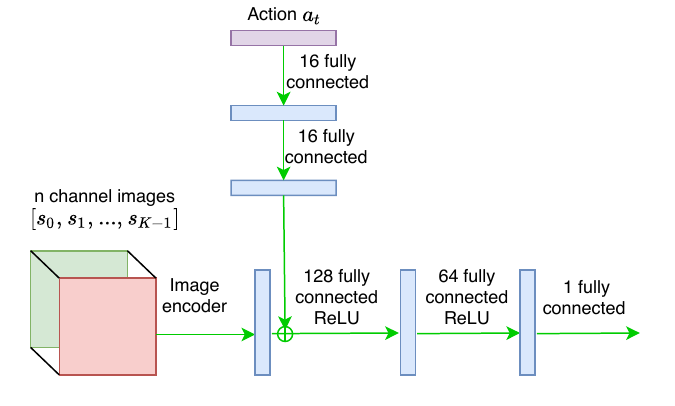}\label{fig:q_network}}
\caption{Network architecture}
\end{figure*}

The agent receives a $64 \times 36$ image from the 3D navigation environment every time step. The image encoder shown in Figure~\ref{fig:image_encoder} converts the last $K$ images the agent receives from the environment into the corresponding state vector. Each image is passed through a series of 2D Convolution networks with ReLU activation, then the output is flattened to a vector and stacked with all the $K$ time step outputs $[s_0, s_1, ..., s_{K-1}]$. The 2D convolution networks use shared weights and were implemented using a single network parallelly processing the $n$ channel images. In the scenario of simple navigation the input is grayscale with $n=1$, and for the color environment it is an RGB image with $n=3$. The input to the series of $1D$ convolution networks is of shape $(K, 64)$ and the resulting output is a $256$ dimensional vector, which is the current state vector that is the input to the policy and Q networks. 

The policy returns $(\mu, \sigma)$ and action $a_t$ is obtained by taking the hyperbolic tangent: $tanh(x)$, where $x \sim \mathcal{N}(\mu, \sigma^2)$ as shown in Figure~\ref{fig:policy_network}. The output action in the $3D$ navigation environments is a $1$ dimensional continuous action in $[-1, 1]$ corresponding to the angle of the steering wheel of the simulated vehicle for the specific time step $t$, and is fed into the environment as $30 \times a_t$. During evaluation, the mean of the distribution is chosen as the action. The $Q$ network architecture as shown in Figure~\ref{fig:q_network} is almost identical to the policy network, except it also concatenates $a_t$ along with the state vector obtained from the image encoder, the action passed through some linear layers with ReLU activation before concatenation. The termination function  $\beta$ has the same architecture as that of the $Q$ value networks followed by a sigmoid function. For the Riverraid environment we use the exact same architecture as in the DQN~\cite{mnih2013playing} with identical policy and $Q$ networks. Since this a discrete action space, for the purposes of training the network, we make use of the soft actor critic for discrete action setting \cite{christodoulou2019soft}.

The observation for Duckietown is in the form of an RGB image of size $160\times120\times3$. We preprocess this image by converting it into grayscale and cropping it to size $90\times160$. Image encoder is modified considering input size. Here kernel size for $3^{rd}$ 2D conv layer changed to 2 from 3 and seven 1D conv layers are considered by adding 1 more layer. The action is a 1-dimensional continuous action corresponding to the direction of steering. The final module of our policy network is the \textit{tanh} activation function, whose output is scaled by $5$ and given as input action to the environment. For this environment, even during evaluation after the options are all trained, the actions are sampled from the normal distribution parameterized by the outputs of the policy network.

For the implementation of the soft option critic, we made use of shared networks with an additional parameter $o_t$, which is the one hot vector representation of the option selected at the time. As with action $a_t$, the one hot vector $o_t$ is also passed through a similar network before being concatenated with the state vector.

All our experiments make use of the Adam optimizer and all buffers used for updating the termination functions via binary cross entropy loss, containing roll-out trajectories after a policy has been trained, have a maximum size of $1000$. The hyperparameters used for experiments in the 3D navigation environment are given in Table \ref{tab:hyperparams}. The hyperparameters used for the other settings, the 3D color navigation environment, the Atari Riverraid environment and the Duckietown environment, are provided in Tables \ref{tab:table2}, \ref{tab:table3} and \ref{tab:table4} respectively. In the 3D navigation environment, only the first policy's termination function is not updated using binary cross entropy as mentioned in the paper, as it tends to generalize rather well. The hyperparameters for the goal based settings are given in Tables \ref{tab:table5} and \ref{tab:table6}.

\begin{table}
    \caption{3D Color Navigation environment}
    \label{tab:table2}
    \centering
    \begin{tabular}{ll} 
      \hline
      \textbf{Parameter} & \textbf{Value}\\
      \hline
      learning rate & $3 \times 10^{-4}$\\
      discount factor($\gamma$) & $0.99$\\
      replay buffer size & $10^4$\\
      target smoothing coefficient($\tau$) & $0.005$\\
      number of frames in input state ($K$) & $10$ \\
      batch size & $16$ \\
      alpha threshold ($\alpha_{min}$) & $0.01$ \\
      termination discount factor ($\gamma_\beta$) & $0.9$\\
      \hline
    \end{tabular}
\end{table}

\begin{table}
  
    \caption{Atari Riverraid environment}
    \label{tab:table3}
    \centering
    \begin{tabular}{ll} 
      \hline
      \textbf{Parameter} & \textbf{Value}\\
      \hline
      learning rate & $3 \times 10^{-4}$\\
      discount factor($\gamma$) & $0.99$\\
      replay buffer size & $10^4$\\
      target smoothing coefficient($\tau$) & $0.005$\\
      number of frames in input state ($K$) & $4$ \\
      batch size & $16$ \\
      alpha threshold ($\alpha_{min}$) & $0.01$ \\
      termination discount factor ($\gamma_\beta$) & $0.95$\\
      \hline
    \end{tabular}
\end{table}

\begin{table}[t]
    \caption{Duckie Town environment}
    \label{tab:table4}
    \centering
    \begin{tabular}{ll} 
      \hline
      \textbf{Parameter} & \textbf{Value}\\
      \hline
      learning rate & $3 \times 10^{-4}$\\
      discount factor($\gamma$) & $0.99$\\
      replay buffer size & $10^3$\\
      target smoothing coefficient($\tau$) & $0.005$\\
      number of frames in input state ($K$) & $4$ \\
      batch size & $16$ \\
      alpha threshold ($\alpha_{min}$) & $0.01$ \\
      termination discount factor ($\gamma_\beta$) & $0.95$\\
      \hline
    \end{tabular}
\end{table}

\begin{table}
    \caption{3D environment with arrow instruction}
    \label{tab:table5}
    \centering
    \begin{tabular}{ll} 
      \hline
      \textbf{Parameter} & \textbf{Value}\\
      \hline
      learning rate & $3 \times 10^{-4}$\\
      discount factor($\gamma$) & $0.99$\\
      replay buffer size & $10^3$\\
      target smoothing coefficient($\tau$) & $0.005$\\
      number of frames in input state ($K$) & $4$ \\
      batch size & $16$ \\
      alpha threshold ($\alpha_{min}$) & $0.01$ \\
      termination discount factor ($\gamma_\beta$) & $0.95$\\
      \hline
    \end{tabular}
\end{table}

\begin{table}
    \caption{Duckie Town environment with arrow instruction}
    \label{tab:table6}
    \centering
    \begin{tabular}{ll} 
      \hline
      \textbf{Parameter} & \textbf{Value}\\
      \hline
      learning rate & $3 \times 10^{-4}$\\
      discount factor($\gamma$) & $0.99$\\
      replay buffer size & $10^3$\\
      target smoothing coefficient($\tau$) & $0.005$\\
      number of frames in input state ($K$) & $6$ \\
      batch size & $16$ \\
      alpha threshold ($\alpha_{min}$) & $0.01$ \\
      termination discount factor ($\gamma_\beta$) & $0.97$\\
      \hline
    \end{tabular}
\end{table}

\end{document}